\documentclass[sn-mathphys,Numbered]{sn-jnl}

\usepackage{graphicx}%
\usepackage{multirow}%
\usepackage{amsmath,amssymb,amsfonts}%
\usepackage{amsthm}%
\usepackage{mathrsfs}%
\usepackage[title]{appendix}%
\usepackage{xcolor}%
\usepackage{textcomp}%
\usepackage{manyfoot}%
\usepackage{booktabs}%
\usepackage{algorithm}%
\usepackage{algorithmicx}%
\usepackage{algpseudocode}%
\usepackage{listings}%

\usepackage{pdfpages}
\usepackage{tikz}
\newcommand{\argmax}{\operatornamewithlimits{argmax}}

\usepackage{amsmath}
\usepackage{amssymb}
\usepackage{mathtools}

\theoremstyle{thmstyleone}%
\theoremstyle{thmstyletwo}%

\theoremstyle{thmstylethree}%

\begin{document}

\title[Feature Importance Depends on Properties of the Data]{Feature Importance Depends on Properties of the Data: Towards Choosing the Correct Explanations for Your Data and Decision Trees based Models}


\author*[1,2]{\fnm{Célia Wafa} \sur{AYAD}}\email{wafa.ayad@polytechnique.edu}

\author[2]{\fnm{Thomas} \sur{Bonnier}}\email{thomas.bonnier@socgen.com}

\author[2]{\fnm{Benjamin} \sur{Bosch}}\email{benjamin.bosch@socgen.com}

\author[3]{\fnm{Sonali} \sur{Parbhoo}}\email{s.parbhoo@imperial.ac.uk}
\author[1]{\fnm{Jesse} \sur{Read}}\email{jesse.read@polytechnique.edu}

\affil*[1]{\orgdiv{LIX}, \orgname{Ecole polytechnique, IP Paris}, \orgaddress{ \state{Palaiseau}, \country{France}}}

\affil[2]{\orgdiv{MRM}, \orgname{Société Générale}, \orgaddress{\state{Paris}, \country{France}}}

\affil[3]{\orgdiv{Department of Engineering}, \orgname{Imperial College London}, \orgaddress{\state{London}, \country{UK}}}


\abstract{In order to ensure the reliability of the explanations of machine learning models, it is crucial to establish their advantages and limits and in which case each of these methods outperform. However, the current understanding of when and how each method of explanation can be used is insufficient. To fill this gap, we perform a comprehensive empirical evaluation by synthesizing multiple datasets with the desired properties.
Our main objective is to assess the quality of feature importance estimates provided by local explanation methods, which are used to explain predictions made by decision tree-based models. By analyzing the results obtained from synthetic datasets as well as publicly available binary classification datasets, we observe notable disparities in the magnitude and sign of the feature importance estimates generated by these methods. Moreover, we find that these estimates are sensitive to specific properties present in the data.
Although some model hyper-parameters do not significantly influence feature importance assignment, it is important to recognize that each method of explanation has limitations in specific contexts. Our assessment highlights these limitations and provides valuable insight into the suitability and reliability of different explanatory methods in various scenarios.}




\maketitle

\section{Introduction}
Decision tree-based models such as random forest \cite{breimanRandomForests2001a} are widely used machine learning algorithms in data science. Although deep learning has been increasingly popular, especially in domains such as computer vision and natural language processing, random forest, for example, continues to be a competitive option in many types of tabular data in a diverse number of domains, including biology \cite{Multitree} and medicine \cite{ricordeauApplicationRandomForests2010}, where interpretation is paramount. 
Small decision trees operating on understandable feature spaces are naturally interpretable, and although this interpretability is diluted across a large forest, it can be recovered in terms of feature importance, which is a major tool that can be used in practical applications for data understanding, model improvement, or model explainability. 
However, practitioners may lose trust in the importance scores provided for random forest \cite{stroblBiasRandomForest2007}, or simply be unable to use them to answer their research questions from the feature importance result due to a number of reasons \cite{BewareDefaultRandom,LimitationsInterpretableMachine}, for example:
(1) a relative lack of training examples leads to instability where the importance scores change due to only minor changes or additions to the dataset or hyperparameters. (2) Even with a large training set, multiple (possibly equivalent) feature scores can be presented. (3) the feature importance scoring mechanism is thrown off by particular properties of the data distribution such as noise, imbalance, and feature type (in particular, the importance of continuous features is often overestimated).
(4) results where the importance of the feature is assigned to spurious or even random features.  
Practitioners are thus often right to be reluctant to draw conclusions from or place trust in off-the-shelf feature-importance scorers, and we aim to remedy this to some extent with a benchmarking study. 

To remedy this, researchers proposed explainability methods such as \textsf{LIME} \cite{ribeiroWhyShouldTrust2016} and \textsf{SHAP} \cite{lundbergUnifiedApproachInterpreting2017} to explain black-box models by attributing feature importance estimates as explanations of the model's predictions. 
While prior research 
\cite{krishnaDisagreementProblemExplainable2022,attanasioFerretFrameworkBenchmarking2022,camburuCanTrustExplainer2019,bodriaBenchmarkingSurveyExplanation2021,neelyOrderCourtExplainable2021} has already taken the first steps towards analyzing the disagreement of explanation methods for models such as deep neural networks, analyzing the behavior of the wide range of existing explanation methods for random forest or in general ensemble trees still insufficiently explored, with regard to particular data properties and model parameters \cite{flora2022comparing}.

Compared to other work, we study the explainability methods suited to explain decision tree-based models. Some of these methods are specific to tree ensembles and the rest are general model-agnostic (which, thus, can also be applied to random forest). We do so with extensive experiments on synthetic alongside real-world datasets, and certain manipulations thereof, which we carry out to isolate and identify aspects which lead to particular results insofar as feature importance. This provides a more thorough understanding, which we use to highlight some limitations of existing methods, and formulate a number of recommendations for practitioners. 
The contribution in this paper is twofold. 
\begin{itemize}
    \item Conduct a thorough evaluation of various explainability methods in the context of specific data properties, such as noise levels, feature correlations, and class imbalance, elucidating their strengths and limitations.
    \item Providing valuable guidance for practitioners and researchers on selecting the most suitable explainability method based on the characteristics of their dataset.
    
\end{itemize}

\section{Background}
\paragraph{Post-hoc local explanations for tree based models}\label{meths}
\begin{table}
\centering
	\begin{tabular}{|l|l|}
		\hline
		 Acronym  & Method\\
		\hline
           \textsf{LSurro} & Local surrogates \cite{molnar2022} \\
           \textsf{LIME} & Local surrogate models \cite{ribeiroWhyShouldTrust2016}\\
           \textsf{Kshap} &  Kernel  \textsf{SHAP} \cite{lundbergUnifiedApproachInterpreting2017}\\
           Sshap&  Sampling  \textsf{SHAP} \cite{lundbergUnifiedApproachInterpreting2017}\\
           \textsf{Tshap} & Tree  \textsf{SHAP} \cite{lundbergConsistentIndividualizedFeature2019a}\\
           \textsf{TI} & Tree Interpreter \cite{treeinterpreter}\\
		 \hline
	\end{tabular}
 \caption{\label{tab:accr} Explainability methods under consideration.}
\end{table}
Post-hoc explanation methods can be classified based on explanation scope (global vs. local), model architecture (specific vs. agnostic) and basic unit of explanation (feature importance vs. rule-based). This paper focuses on local post-hoc explanation methods based on feature importance. It analyzes four model agnostic methods in Table \ref{tab:accr} (LIME, \textsf{LSurro}, \textsf{Kshap} and Sshap) and two tree-specific methods (\textsf{Tshap} and TI).

LIME, \textsf{LSurro}, \textsf{Kshap} and \textsf{Sshap} construct local interpretable models such as a linear regression in the neighborhood of the instance that is being explained. These methods differ in the kernel used for the generation of the local neighborhood and the objective that is being optimized. For example, \textsf{LIME} uses an exponential kernel while  \textsf{SHAP} explainers use Shapley kernel, and both use the squared error to minimize the loss between the black box model and the surrogate model.
On the other hand, \textsf{Tshap} and \textsf{TI} both are designed to explain tree based methods such as random forest. \textsf{Tshap} computes the Shapley value for each node in the tree based on the decision splits. While \textsf{TI} decomposes the prediction into the sum of feature contributions and the bias, i.e; the mean given by the root of the decision tree that covers the entire training set.

\paragraph{Properties (metrics) of local explanations} \label{metrics}
Since we cannot measure the accuracy of feature importance estimates due to the absence of ground truth feature importance, prior research proposed objectively assessing the quality of explanations through various metrics, such as examining stability and compactness of the local explanations, the faithfulness of the interpretable model to the black-box predictions \cite{alvarez-melisRobustnessInterpretabilityMethods2018}, robustness to input perturbations \cite{bodriaBenchmarkingSurveyExplanation2021} and fairness across subgroups \cite{rajbahadurImpactFeatureImportance2022}. 
The stability metric \cite{alvarez-melisRobustnessInterpretabilityMethods2018} compares the explanation given to an instance in its neighborhood. If two instances have similar feature values and predictions, they should have similar explanations. 
While the compactness metric shows whether it is possible to explain the model's prediction with fewer features, so that it is easily understood by humans.
To measure the robustness of the interpreters, local fidelity and stability were also proposed \cite{slackReliablePostHoc2021,petsiukRISERandomizedInput2018}. The faithfulness can also be measured by the consistency, the feature and rank agreements \cite{krishnaDisagreementProblemExplainable2022} between the explanation and the ground truth feature importance or between pairs of explanations generated by different methods. 

\section{Related Work To explainability Benchmarking Frameworks}

The landscape of explainable artificial intelligence has witnessed a surge in research efforts aimed at understanding and evaluating the diverse methodologies employed for interpreting complex machine learning models. Several survey and benchmarking papers, including XAI-survey \cite{bodriaBenchmarkingSurveyExplanation2023a} and BenchXAI \cite{liuSyntheticBenchmarksScientific2021, liu2021synthetic}, have played a crucial role in shedding light on the disagreement problem within existing explainability methods \cite{krishnaDisagreementProblemExplainable2022, neelyOrderCourtExplainable2021, camburuCanTrustExplainer2019, hanWhichExplanationShould2022, turbeEvaluationPosthocInterpretability2023}. Notably, these contributions have been important to the understanding of the challenges and nuances associated within the field of machine learning explainability.

While the majority of existing benchmarks have primarily focused on explaining neural networks for text and image data with feature importance generation methods such as \cite{ismailBenchmarkingDeepLearning2020, attanasioFerretFrameworkBenchmarking2022, bodriaBenchmarkingSurveyExplanation2021, yang2019benchmarking, zhong2023clock, han2022explanation}, 
the research community has introduced several frameworks to facilitate the transparent evaluation of explainability methods. Examples include OpenXAI \cite{agarwalOpenXAITransparentEvaluation}, Captum \cite{kokhlikyanCaptumUnifiedGeneric2020}, Quantus \cite{hedstromQuantusExplainableAI2022}, and many others such as \cite{guidotti2021evaluating, le2023benchmarking}.
In addition, \cite{turbe2023evaluation} introduced a quantitative framework with specific metrics for assessing the performance of post-hoc interpretability methods, particularly in the context of time-series classification. This research provides a targeted approach to evaluating the temporal aspects of interpretability.
These frameworks aim to provide a structured approach to assess the effectiveness and reliability of various explainability techniques. 

Despite these advancements, the evaluation of post-hoc interpretability methods for ensemble trees predictions, taking into account different data properties, remains unexplored. This paper seeks to fill this gap by addressing the specific question of how existing interpretability methods designed for ensemble trees predictions perform under varying data conditions. This research aims to contribute valuable insights and further enrich the evolving field of explainability evaluation.
\section{Synthetic Data Generation Framework}
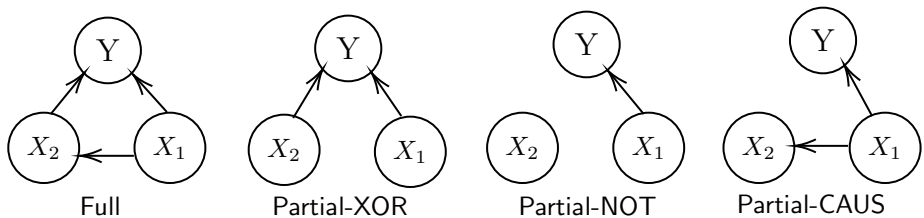
\begin{figure}[h]
	\centering
\tikzset{every picture/.style={line width=0.75pt}} 

\begin{tikzpicture}[x=0.75pt,y=0.75pt,yscale=-1,xscale=1]

\draw    (353,187) -- (366.97,163.71) ;
\draw [shift={(368,162)}, rotate = 120.96] [color={rgb, 255:red, 0; green, 0; blue, 0 }  ][line width=0.75]    (10.93,-3.29) .. controls (6.95,-1.4) and (3.31,-0.3) .. (0,0) .. controls (3.31,0.3) and (6.95,1.4) .. (10.93,3.29)   ;
\draw    (407,186) -- (393.13,165.65) ;
\draw [shift={(392,164)}, rotate = 55.71] [color={rgb, 255:red, 0; green, 0; blue, 0 }  ][line width=0.75]    (10.93,-3.29) .. controls (6.95,-1.4) and (3.31,-0.3) .. (0,0) .. controls (3.31,0.3) and (6.95,1.4) .. (10.93,3.29)   ;
\draw    (232,185) -- (247.74,165.55) ;
\draw [shift={(249,164)}, rotate = 128.99] [color={rgb, 255:red, 0; green, 0; blue, 0 }  ][line width=0.75]    (10.93,-3.29) .. controls (6.95,-1.4) and (3.31,-0.3) .. (0,0) .. controls (3.31,0.3) and (6.95,1.4) .. (10.93,3.29)   ;
\draw    (290,184) -- (274.33,166.49) ;
\draw [shift={(273,165)}, rotate = 48.18] [color={rgb, 255:red, 0; green, 0; blue, 0 }  ][line width=0.75]    (10.93,-3.29) .. controls (6.95,-1.4) and (3.31,-0.3) .. (0,0) .. controls (3.31,0.3) and (6.95,1.4) .. (10.93,3.29)   ;
\draw    (273.5,206) -- (248.5,206) ;
\draw [shift={(246.5,206)}, rotate = 360] [color={rgb, 255:red, 0; green, 0; blue, 0 }  ][line width=0.75]    (10.93,-3.29) .. controls (6.95,-1.4) and (3.31,-0.3) .. (0,0) .. controls (3.31,0.3) and (6.95,1.4) .. (10.93,3.29)   ;
\draw    (528,185) -- (511.27,164.55) ;
\draw [shift={(510,163)}, rotate = 50.71] [color={rgb, 255:red, 0; green, 0; blue, 0 }  ][line width=0.75]    (10.93,-3.29) .. controls (6.95,-1.4) and (3.31,-0.3) .. (0,0) .. controls (3.31,0.3) and (6.95,1.4) .. (10.93,3.29)   ;
\draw    (642,186) -- (628.95,161.76) ;
\draw [shift={(628,160)}, rotate = 61.7] [color={rgb, 255:red, 0; green, 0; blue, 0 }  ][line width=0.75]    (10.93,-3.29) .. controls (6.95,-1.4) and (3.31,-0.3) .. (0,0) .. controls (3.31,0.3) and (6.95,1.4) .. (10.93,3.29)   ;
\draw    (630,201) -- (605,201) ;
\draw [shift={(603,201)}, rotate = 360] [color={rgb, 255:red, 0; green, 0; blue, 0 }  ][line width=0.75]    (10.93,-3.29) .. controls (6.95,-1.4) and (3.31,-0.3) .. (0,0) .. controls (3.31,0.3) and (6.95,1.4) .. (10.93,3.29)   ;

\draw    (530.5, 202) circle [x radius= 17.69, y radius= 17.69]   ;
\draw (530.5,202) node    {$X_{1}$};
\draw    (467.5, 202) circle [x radius= 17.69, y radius= 17.69]   ;
\draw (467.5,202) node    {$X_{2}$};
\draw    (499.5, 150) circle [x radius= 16.51, y radius= 16.51]   ;
\draw (499.5,150) node  [font=\large]  {$\text{Y}$};
\draw    (411.5, 204) circle [x radius= 17.69, y radius= 17.69]   ;
\draw (411.5,204) node    {$X_{1}$};
\draw    (348.5, 203) circle [x radius= 17.69, y radius= 17.69]   ;
\draw (348.5,203) node    {$X_{2}$};
\draw    (380.5, 152) circle [x radius= 16.51, y radius= 16.51]   ;
\draw (380.5,152) node  [font=\large]  {$\text{Y}$};
\draw    (648.5, 202) circle [x radius= 17.69, y radius= 17.69]   ;
\draw (648.5,202) node    {$X_{1}$};
\draw    (585.5, 202) circle [x radius= 17.69, y radius= 17.69]   ;
\draw (585.5,202) node    {$X_{2}$};
\draw    (617.5, 148) circle [x radius= 16.51, y radius= 16.51]   ;
\draw (617.5,148) node  [font=\large]  {$\text{Y}$};
\draw    (291.5, 202) circle [x radius= 17.69, y radius= 17.69]   ;
\draw (291.5,202) node    {$X_{1}$};
\draw    (228.5, 202) circle [x radius= 17.69, y radius= 17.69]   ;
\draw (228.5,202) node    {$X_{2}$};
\draw    (260.5, 153) circle [x radius= 16.51, y radius= 16.51]   ;
\draw (260.5,153) node  [font=\large]  {$\text{Y}$};
\draw (245,225) node [anchor=north west][inner sep=0.75pt]   [align=left] {\textsf{Full}};
\draw (341,225) node [anchor=north west][inner sep=0.75pt]   [align=left] {\textsf{Partial-XOR}};
\draw (464,225) node [anchor=north west][inner sep=0.75pt]   [align=left] {\textsf{Partial-NOT}};
\draw (573,224) node [anchor=north west][inner sep=0.75pt]   [align=left] {\textsf{Partial-CAUS}};

\end{tikzpicture}
\caption{\label{fig:bn}Bayesian networks that we use as a schema to generate synthetic data, illustrating one full and three partial factorizations of $P(X,Y)$. 
}
\end{figure}
We wish that the decision function of a model $f$ make predictions $\hat y = f(X)$ to minimize some expected loss; \[\mathbb{E}_{X,Y \sim P(Y,X)}[\ell(Y,f(X))]\] where $P(Y,X)$ is the data distribution and $\ell$ is some loss function. A decision function $f : \mathcal{X} \rightarrow \mathcal{Y}$ is essentially some function of $P(X,Y)$. For example, to maximize classification accuracy,
   \[
		y =  f(X) = 
		 \argmax_y P(y|X)
   \]
Therefore, the feature importance outcome depends significantly on different mechanisms/properties of this generating distribution $P(X,Y)$. Namely, the properties of each variable (and its distribution), and the properties of the relations among variables. 
Considering only two features, we could represent the concept as a Bayesian network (Fig.~\ref{fig:bn} illustrates). 
\[
    P(X,Y) = P(X_1,X_2,Y) = P(Y|X_1,X_2)P(X_2|X_1)P(X_1)
\]
i.e., a Full factorization of the joint, and thus we could consider the following properties (as nodes and edges): 
\begin{enumerate}
    \item $P(X_1)$: specifying the type of the feature $X_1$; 
    \item $P(X_2|X_1)$: the amount of conditional dependence of $X_2$ on $X_1$; and
	\item $P(Y|X_1,X_2)$: the amount and type of correlation between features and target, revealing the special case of $P(Y|X_1,X_2)=P(Y)$ when there is no correlation.
	
\end{enumerate}
\textsf{Partial-XOR} and \textsf{Partial-NOT} exhibit feature independence (features are independent from each other -- when the target is observed), and both features are required to make a perfect prediction for \textsf{Partial-XOR}, and only $X_1$ is required to perfectly predict \textsf{Partial-NOT}; in this case deterministic.

In real-world settings,  \textsf{Full} represents the case where one feature is related to another feature and both participate to make the prediction of the output, for example in \textsc{Adult Income} dataset (that we later include in our experiments) the feature \textsc{occupation} is correlated to \textsc{age} and both predict the output \textsc{income}. \textsf{Partial-CAUS} on the other hand, may represent a causal relationship between one feature and the outcome through a chain of causality; or an indirect correlation of one feature on the output through another feature(or even multiple features), forming a chain of correlations.

\textsf{Full} and \textsf{Partial-CAUS} are specific cases of respectively \textsf{Partial-XOR} and \textsf{Partial-NOT} when $X_1$ is dependent on $X_2$, thus for the rest of this paper, we denote \textsf{XOR} to refer to both \textsf{Full} and \textsf{Partial-XOR}, and use \textsf{NOT} to refer to \textsf{Partial-CAUS} and \textsf{Partial-NOT}.
The data can be generated as:

\[
    P(X) \sim \mathcal{N}(\mu, \Sigma) 
\]

Where $\mathcal{N}$ is a bi-variate normal distribution and $\Sigma$ is the covariance matrix, representing the amount of correlation between the two random variables $X_1$ and $X_2$.
In order to introduce noise, we apply a mask to invert a percentage of labels in the output $y$. Let's denote the percentage of labels to invert as $\epsilon$. Let $M_i$ be a binary mask defined as:
\[ M_i = 
\begin{cases} \label{eq1}
1 & \text{with probability } \epsilon \\
0 & \text{with probability } 1-\epsilon
\end{cases}
\]
The perturbed output $y$ can be obtained with :
\[ \hat{y} = y \odot (1 - M) + (1 - y) \odot M \]
With $y=f(X)$ and $f$ represents the logical operators \textsf{XOR} or \textsf{NOT}.


\paragraph{Ground truth feature importance}\label{gtfi}
We use $\phi^*_{X}(f^*)$ to denote the ground truth feature importance that are given by the true model $f^*$ to which we compare $\phi_{X}(f)$, the feature importance estimates that is generated by each of the local explainability methods to explain the predictions of the learned model $f$.
Intuitively, the true model $f^*$ can be illustrated with a $D$-depth decision tree. With $D=2$ for \textsf{XOR} dataset variants (the first split on $X_1$ and the second on $X_2$) and $D=1$  for \textsf{NOT} dataset variants (only one split on $X_1$).

When $\epsilon = 0$, the ground truth feature importance $\phi^{\star}_{X}$ for all variants of \textsf{XOR} datasets are fixed as $\phi^{\star}_{X_1}$ = $\phi^{\star}_{X_2}$ = .5, because both $X_1$ and $X_2$ are \emph{necessary} to make the prediction of \textsf{XOR}. The amount of the correlation $\rho$ between $X_1$ and $X_2$ doesn't affect the importance as both are \emph{necessary} to make the prediction of \textsf{XOR}. Meanwhile, only $X_1$ is \emph{necessary} to make the prediction of \textsf{NOT}, thus $\phi^{\star}_{X_1}$ = 1 and $\phi^{\star}_{X_2}$ = $\rho$, because when $X_2$ is correlated to $X_1$, $X_2$ have an indirect influence estimated by $\rho$ to predict \textsf{NOT}. 

On the other hand, when $\epsilon \neq 0$, $\phi^{\star}_{X_1} = \phi^{\star}_{X_2} = .5 * \epsilon$ for \textsf{XOR} dataset variants, and $\phi^{\star}_{X_1} = 1 * \epsilon$ and $\phi^{\star}_{X_2} = \rho * \epsilon$ for \textsf{NOT} dataset variants.



For the XOR function, both \(x_1\) and \(x_2\) are equally important in determining the output, and their importance scores should ideally converge to \(0.5\) when considering a large amount of data points.
Consider a decision tree model that aims to predict the XOR function using features \(x_1\) and \(x_2\). For simplicity, let's assume that the decision tree splits on both \(x_1\) and \(x_2\) at each level. The decision tree's predictions can be expressed as:
\[ \hat{y} = f(x_1, x_2) \]
Now, let's define the feature importance scores (\(\phi_{x_1}\) and \(\phi_{x_2}\)) using the Gini impurity criterion, a common metric for decision trees:
\[ \phi_{x_1} = \sum_{\text{nodes splitting on } x_1} \text{Gini decrease at the node} \]
\[ \phi_{x_2} = \sum_{\text{nodes splitting on } x_2} \text{Gini decrease at the node} \]
In a large dataset, the decision tree will be able to accurately capture the XOR relationship, and both \(x_1\) and \(x_2\) should contribute equally to the impurity decrease, leading to similar importance scores. For a balanced decision tree, these Gini decreases would be distributed among the splits involving \(x_1\) and \(x_2\). In the limit of a large dataset, we would expect:
\[ \lim_{{\text{large dataset}}} \phi_{x_1} = \lim_{{\text{large dataset}}} \phi_{x_2} = 0.5 \]
This indicates that, as the dataset size increases, the decision tree's feature importance for predicting the XOR function would converge to \(0.5\) for both \(x_1\) and \(x_2\), reflecting their equal importance in determining the output.

\section{Empirical Setup}
To carry out our experiments, we demonstrate our findings on four real-world datasets: \textsc{Heart Diagnosis}, \textsc{Cervical Cancer}, \textsc{Adult Income} and \textsc{German Credit Risk}. These datasets include properties such as feature interactions (dependence or independence),
noise, random irrelevant variables and class imbalance. 

\begin{figure*}
  \centering
  \includegraphics[width=1.1\textwidth]{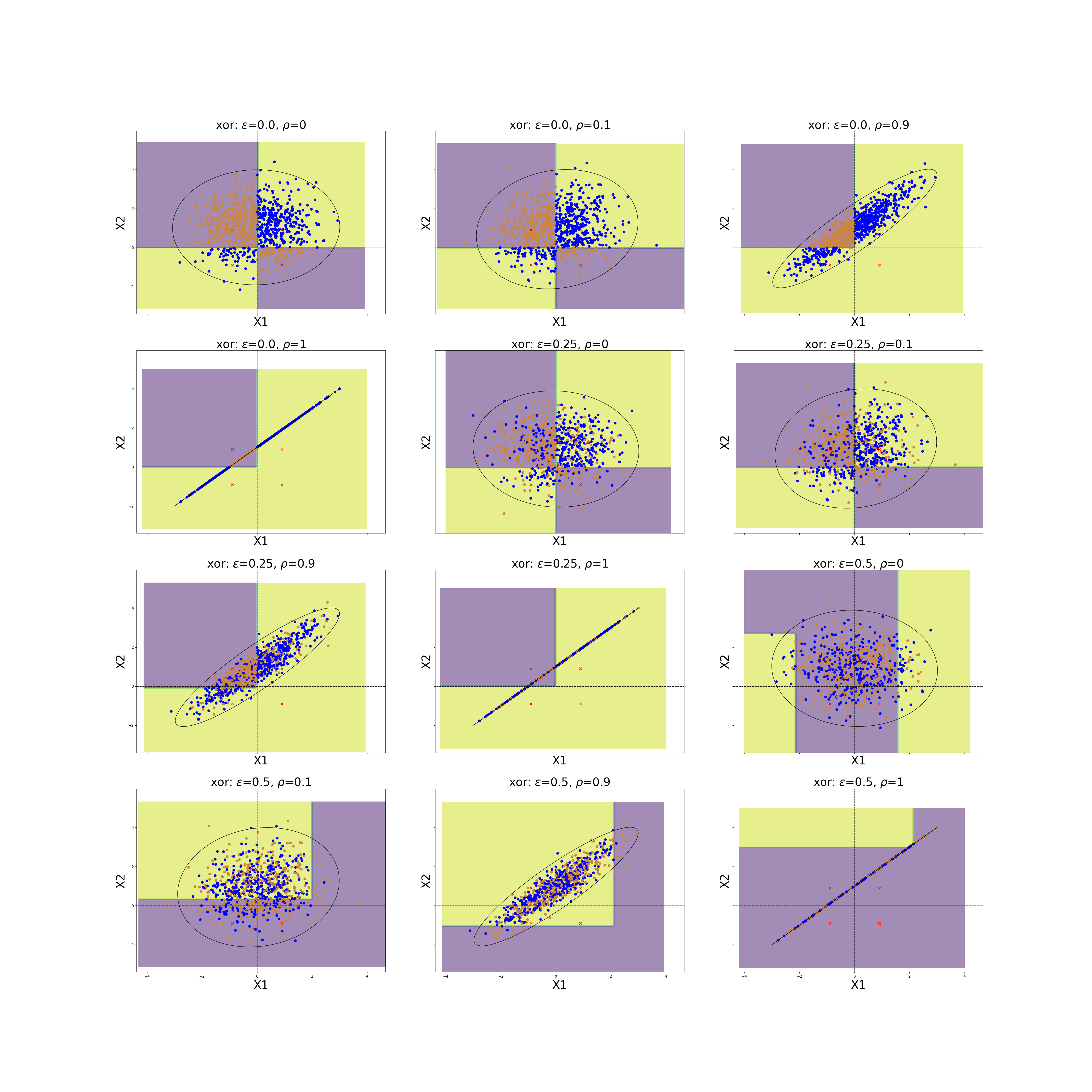}
  \caption{\label{synthdata} Synthetic data XOR. $X \sim \mathcal{N}(\mu, \Sigma)$. Each dataset expresses a different combination of properties. }
\end{figure*}

\begin{figure*}
  \centering
  \includegraphics[width=1.1\textwidth]{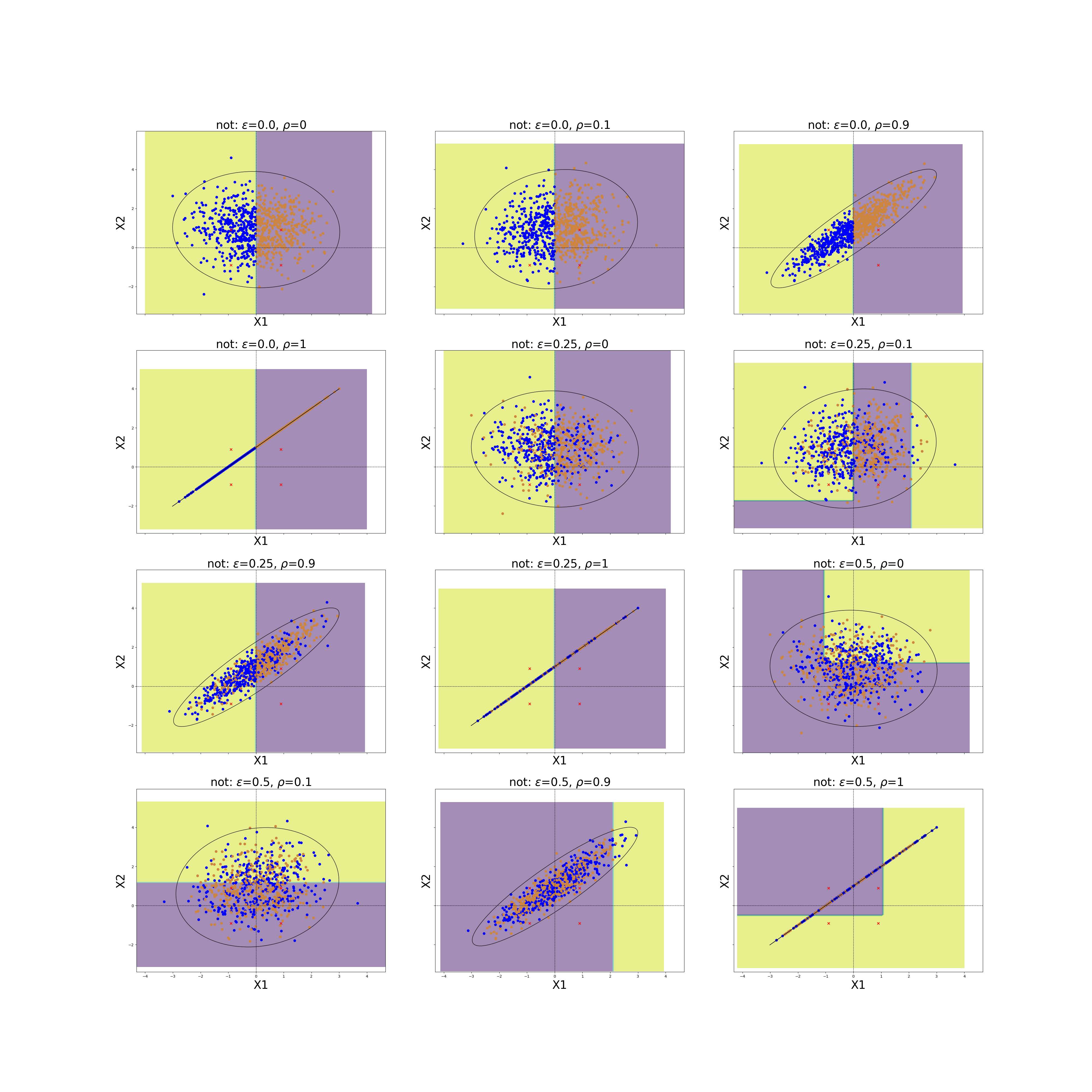}
  \caption{\label{synthdata2} Synthetic data NOT. $X \sim \mathcal{N}(\mu, \Sigma)$. Each dataset expresses a different combination of properties. }
\end{figure*}

We generate 24 synthetic datasets (Figures \ref{synthdata} and \ref{synthdata2}) expressing different combinations of these properties by varying several parameters such as the correlation amount of the normal distribution from which the data points are drawn and the probability of each class, thus, the amount of generated noise and class imbalance. 
Each dataset is divided to 80\% for training and 20\% for testing. We report the results of the feature importance estimates on the test set. 
We compute feature importance estimates on the true model $f^*$ and learned model $f$, so that we can compare the generated feature importance estimates to their ground truth values.
Finally, we analyze the advantages and limitations of each explainability method on the synthetic datasets and we run larger experiments on the above real-world datasets from the UCI repository \cite{Dua:2019}. 


\subsection{Datasets, models and metrics}
\paragraph{Datasets}
Figures \ref{synthdata} and \ref{synthdata2} show the generated datasets by varying the parameters in Eq~\ref{eq1}: 
$\mu \in \{[0,1],[1,0]\}$, $\Sigma \in \{ [[1, 0], [0, 1]], [[1, .1], [.1, 1]] , [[1, .9], [.9, 1]], [[1, 1], [1, 1]] \}$, and $\epsilon \in \{0, .25, .5\}$.

In addition, Table \ref{tab:datasets} summarizes the properties of the four real-world datasets that we use to demonstrate our findings.
\begin{table}[h]
	\centering
    \resizebox{\columnwidth}{!}{
	\begin{tabular}{|l|c|c|c|c|c|}
		\hline
		Dataset & $\#$instances & $\#$features & $\%$ discrete & $\%$ continuous & imbalance \\
		\hline
		  \textsc{Heart Diagnosis} & 303 & 13 & 43 & 57 & yes \\
          \textsc{Cervical Cancer} & 858 & 35 & 62 & 38 & yes \\
          \textsc{Adult Income}& 32561 & 11 & 65 & 35 & no\\
          \textsc{German Credit Risk} & 1\,000 & 23 & 70 & 30 & yes \\ 
		\hline
	\end{tabular}
 }
 \caption{\label{tab:datasets} Summary of the real-world datasets we include in our experiments.}
\end{table}
\paragraph{Models}
For the synthetic datasets, we compute the feature importance scores of 
the learned model $f$ on datasets with 1\,000 instances. The learned model $f$ can be either a decision tree or a random forest.
On the other hand, we use the random forest model with parameters learned using grid search and evaluated with 10-fold cross-validation for each of the real-world datasets. Table \ref{datasetss} summarizes the performances and the feature importance of the decision tree and the random forest models for the generated datasets.
\begin{table}[h]
    \centering
    \resizebox{\columnwidth}{!}{
    \begin{tabular}{|l|c|c|c|c|c|c|}
\hline
Decision function &  $\epsilon$ &  $\rho$ &  DT Accuracy &     DT feature importance &  RF Accuracy &     RF feature importance \\
\hline
              \textsf{XOR} &        0.00 &    0.00 &         1.00 & $\phi_{X_1}$=0.44, $\phi_{X_2}$=0.56 &         0.85 & $\phi_{X_1}$=0.69, $\phi_{X_2}$=0.31 \\
              \textsf{XOR} &        0.00 &    0.10 &         1.00 & $\phi_{X_1}$=0.41, $\phi_{X_2}$=0.59 &         0.90 & $\phi_{X_1}$=0.63, $\phi_{X_2}$=0.37 \\
              \textsf{XOR} &        0.00 &    0.90 &         1.00 & $\phi_{X_1}$=0.51, $\phi_{X_2}$=0.49 &         1.00 & $\phi_{X_1}$=0.55, $\phi_{X_2}$=0.45 \\
              \textsf{XOR} &        0.00 &    1.00 &         1.00 &   $\phi_{X_1}$=0.5, $\phi_{X_2}$=0.5 &         1.00 & $\phi_{X_1}$=0.53, $\phi_{X_2}$=0.47 \\
              \textsf{XOR} &        0.25 &    0.00 &         0.70 & $\phi_{X_1}$=0.47, $\phi_{X_2}$=0.53 &         0.61 & $\phi_{X_1}$=0.62, $\phi_{X_2}$=0.38 \\
              \textsf{XOR} &        0.25 &    0.10 &         0.72 &   $\phi_{X_1}$=0.4, $\phi_{X_2}$=0.6 &         0.62 & $\phi_{X_1}$=0.64, $\phi_{X_2}$=0.36 \\
              \textsf{XOR} &        0.25 &    0.90 &         0.72 & $\phi_{X_1}$=0.51, $\phi_{X_2}$=0.49 &         0.72 & $\phi_{X_1}$=0.56, $\phi_{X_2}$=0.44 \\
              \textsf{XOR} &        0.25 &    1.00 &         0.71 & $\phi_{X_1}$=0.51, $\phi_{X_2}$=0.49 &         0.71 & $\phi_{X_1}$=0.53, $\phi_{X_2}$=0.47 \\
              \textsf{XOR} &        0.50 &    0.00 &         0.52 & $\phi_{X_1}$=0.83, $\phi_{X_2}$=0.17 &         0.50 & $\phi_{X_1}$=0.57, $\phi_{X_2}$=0.43 \\
              \textsf{XOR} &        0.50 &    0.10 &         0.53 & $\phi_{X_1}$=0.69, $\phi_{X_2}$=0.31 &         0.56 & $\phi_{X_1}$=0.57, $\phi_{X_2}$=0.43 \\
              \textsf{XOR} &        0.50 &    0.90 &         0.49 & $\phi_{X_1}$=0.64, $\phi_{X_2}$=0.36 &         0.46 & $\phi_{X_1}$=0.53, $\phi_{X_2}$=0.47 \\
              \textsf{XOR} &        0.50 &    1.00 &         0.54 & $\phi_{X_1}$=0.55, $\phi_{X_2}$=0.45 &         0.47 & $\phi_{X_1}$=0.53, $\phi_{X_2}$=0.47 \\
\hline  
              \textsf{NOT} &        0.00 &    0.00 &         1.00 &   $\phi_{X_1}$=1.0, $\phi_{X_2}$=0.0 &         1.00 & $\phi_{X_1}$=0.84, $\phi_{X_2}$=0.16 \\
              \textsf{NOT} &        0.00 &    0.10 &         1.00 &   $\phi_{X_1}$=1.0, $\phi_{X_2}$=0.0 &         1.00 & $\phi_{X_1}$=0.83, $\phi_{X_2}$=0.17 \\
              \textsf{NOT} &        0.00 &    0.90 &         1.00 &   $\phi_{X_1}$=1.0, $\phi_{X_2}$=0.0 &         1.00 & $\phi_{X_1}$=0.61, $\phi_{X_2}$=0.39 \\
              \textsf{NOT} &        0.00 &    1.00 &         1.00 &   $\phi_{X_1}$=1.0, $\phi_{X_2}$=0.0 &         1.00 & $\phi_{X_1}$=0.49, $\phi_{X_2}$=0.51 \\
              \textsf{NOT} &        0.25 &    0.00 &         0.72 &   $\phi_{X_1}$=1.0, $\phi_{X_2}$=0.0 &         0.72 & $\phi_{X_1}$=0.77, $\phi_{X_2}$=0.23 \\
              \textsf{NOT} &        0.25 &    0.10 &         0.69 & $\phi_{X_1}$=0.99, $\phi_{X_2}$=0.01 &         0.72 &   $\phi_{X_1}$=0.8, $\phi_{X_2}$=0.2 \\
              \textsf{NOT} &        0.25 &    0.90 &         0.71 &   $\phi_{X_1}$=1.0, $\phi_{X_2}$=0.0 &         0.71 & $\phi_{X_1}$=0.63, $\phi_{X_2}$=0.37 \\
              \textsf{NOT} &        0.25 &    1.00 &         0.72 & $\phi_{X_1}$=0.97, $\phi_{X_2}$=0.03 &         0.72 & $\phi_{X_1}$=0.49, $\phi_{X_2}$=0.51 \\
              \textsf{NOT} &        0.50 &    0.00 &         0.50 & $\phi_{X_1}$=0.39, $\phi_{X_2}$=0.61 &         0.48 & $\phi_{X_1}$=0.55, $\phi_{X_2}$=0.45 \\
              \textsf{NOT} &        0.50 &    0.10 &         0.58 &   $\phi_{X_1}$=0.0, $\phi_{X_2}$=1.0 &         0.57 &   $\phi_{X_1}$=0.5, $\phi_{X_2}$=0.5 \\
              \textsf{NOT} &        0.50 &    0.90 &         0.50 & $\phi_{X_1}$=0.69, $\phi_{X_2}$=0.31 &         0.46 & $\phi_{X_1}$=0.53, $\phi_{X_2}$=0.47 \\
              \textsf{NOT} &        0.50 &    1.00 &         0.48 & $\phi_{X_1}$=0.49, $\phi_{X_2}$=0.51 &         0.50 & $\phi_{X_1}$=0.51, $\phi_{X_2}$=0.49 \\
\hline
\end{tabular}
}
    \caption{Parameterization and performances of the decision tree (DT) and the random forest (RF) for the 24 generated datasets with 1.000 instances. Maximum depth of both DT and RF is set to 2.
     }
    \label{datasetss}
\end{table}
\paragraph{Metrics}
To evaluate the quality of the feature importance estimates attributed by the methods in Table \ref{tab:accr}, we compare the feature importance estimates to the ground truth feature importance in Section \ref{gtfi} of the synthetic datasets because the ground truth feature importance estimates in the real-world datasets are hard to obtain. We also evaluate the stability, compactness, consistency, feature and rank agreements for the synthetic and real-world datasets.

\subsection{Experiments}
\subsubsection{Synthetic data}
\begin{figure*}
  \centering
  \includegraphics[width=1.1\textwidth]{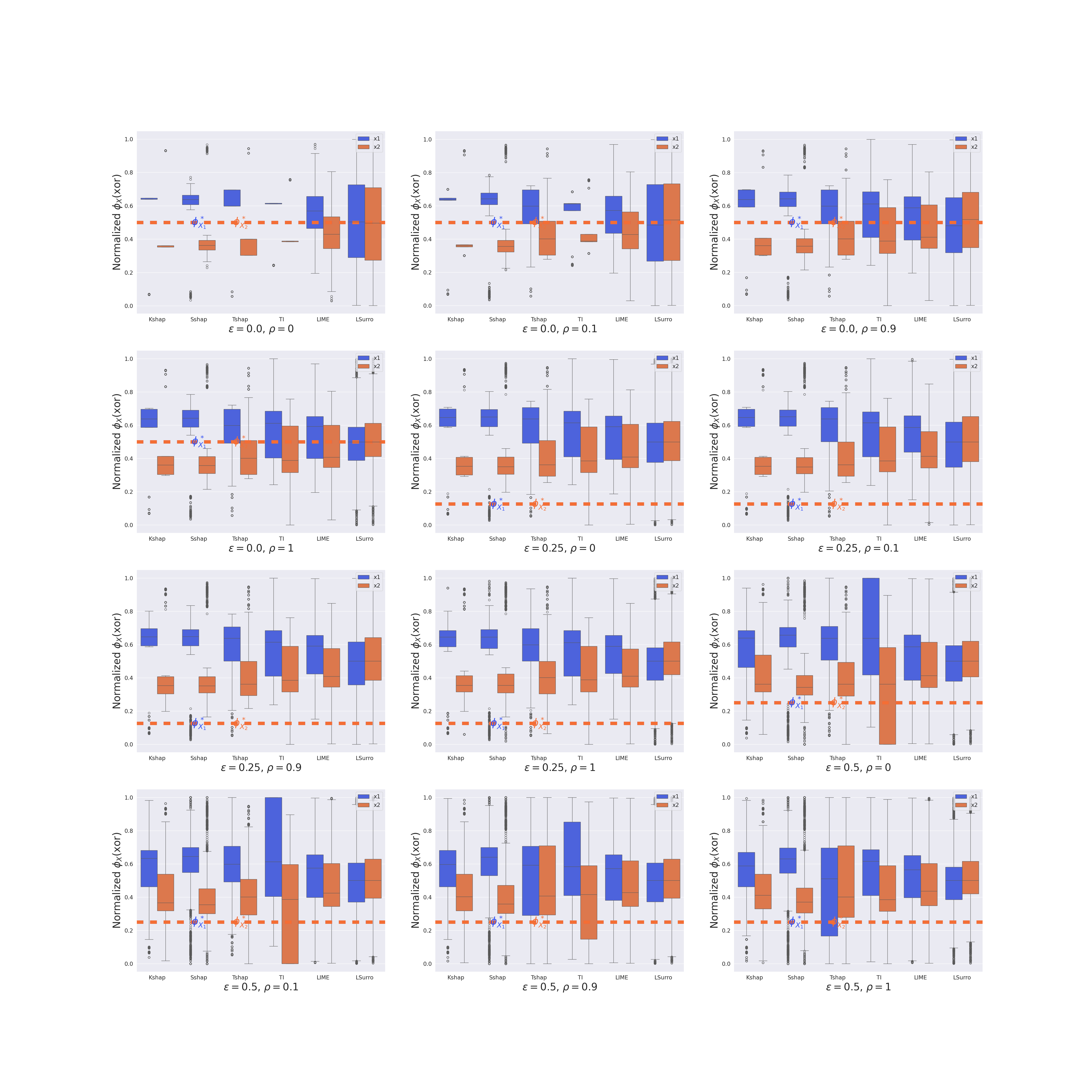}
  \caption{\label{boxplotxor} Normalized feature importance estimates of the \textsf{XOR} datasets. These feature importance estimates are obtained for the decision trees trained on datasets with 1\,000 instances.}
\end{figure*}

\begin{figure*}
  \centering
  \includegraphics[width=1.1\textwidth]{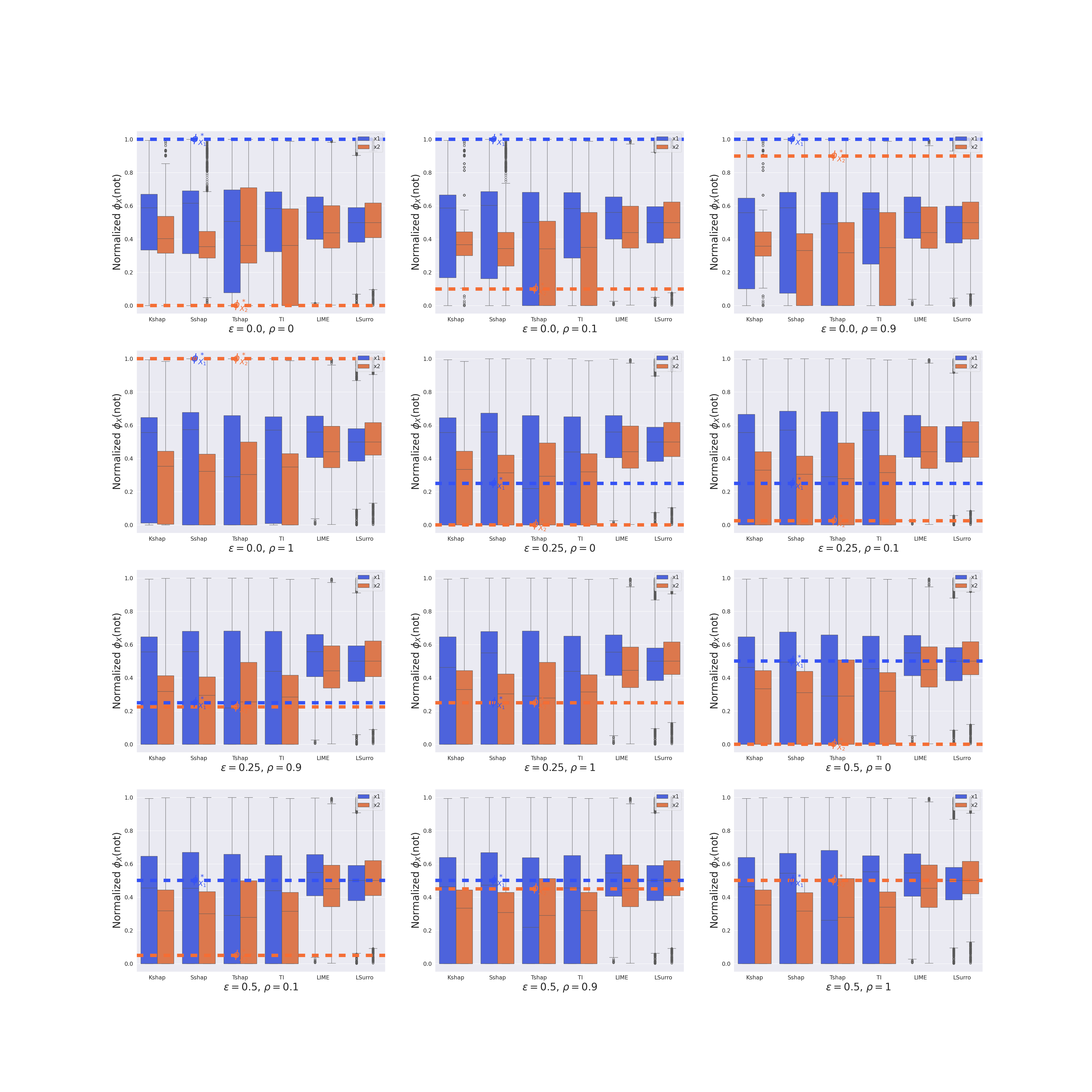}
  \caption{\label{boxplotnot} Normalized feature importance estimates of the \textsf{NOT} datasets. These feature importance estimates are obtained for the decision trees trained on datasets with 1\,000 instances. }
\end{figure*}

Figures \ref{boxplotxor} and \ref{boxplotnot} show the normalized feature importance estimates attributed by the selected explainability methods. After the normalization of the absolute importance of $X_1$ and $X_2$, their contributions sum to one. We perform the normalization to faithfully compare the feature attributions to their ground truth values. 

\paragraph{Explainability methods based on learning surrogate models overestimate the importance to irrelevant variables, Tree interpreter is sensitive to noise and SHAP explainers always favor one feature over the other} 
Overall, all explainers except local surrogates overestimate the importance of $X_1$ over $X_2$ across the \textsf{XOR} datasets. Also, none of these methods perfectly matches ground truth feature importance on average across all datasets.
Moreover, \textsf{LSurro} and \textsf{LIME} feature importance attributions are the least affected by noise and feature correlation. Indeed, \textsf{LSurro} and and \textsf{LIME} attribute comparable importance to $X_1$ and $X_2$ for \textsf{XOR} and \textsf{NOT} dataset variants, and both overestimate the importance of unimportant features (such as $X_2$ in case of \textsf{NOT}). 
Notably, \textsf{TI} is the most affected by noise, that is confirmed in its decomposition of the the feature and noise contributions to the prediction.
Additionally, feature correlations increase the importance and instability of $X_2$ importance in \textsf{XOR} datasets attributed by \textsf{SHAP} explainers, and noise lowers the importance of $X_1$ and $X_2$ for all the explainers.
Finally, \textsf{SHAP} explainers and \textsf{TI} have the highest variance of feature importance estimates in the \textsf{NOT} datasets.


\begin{figure}[h]
  \vspace*{-.5in}
  \centering
   \includegraphics[width=\textwidth/3]{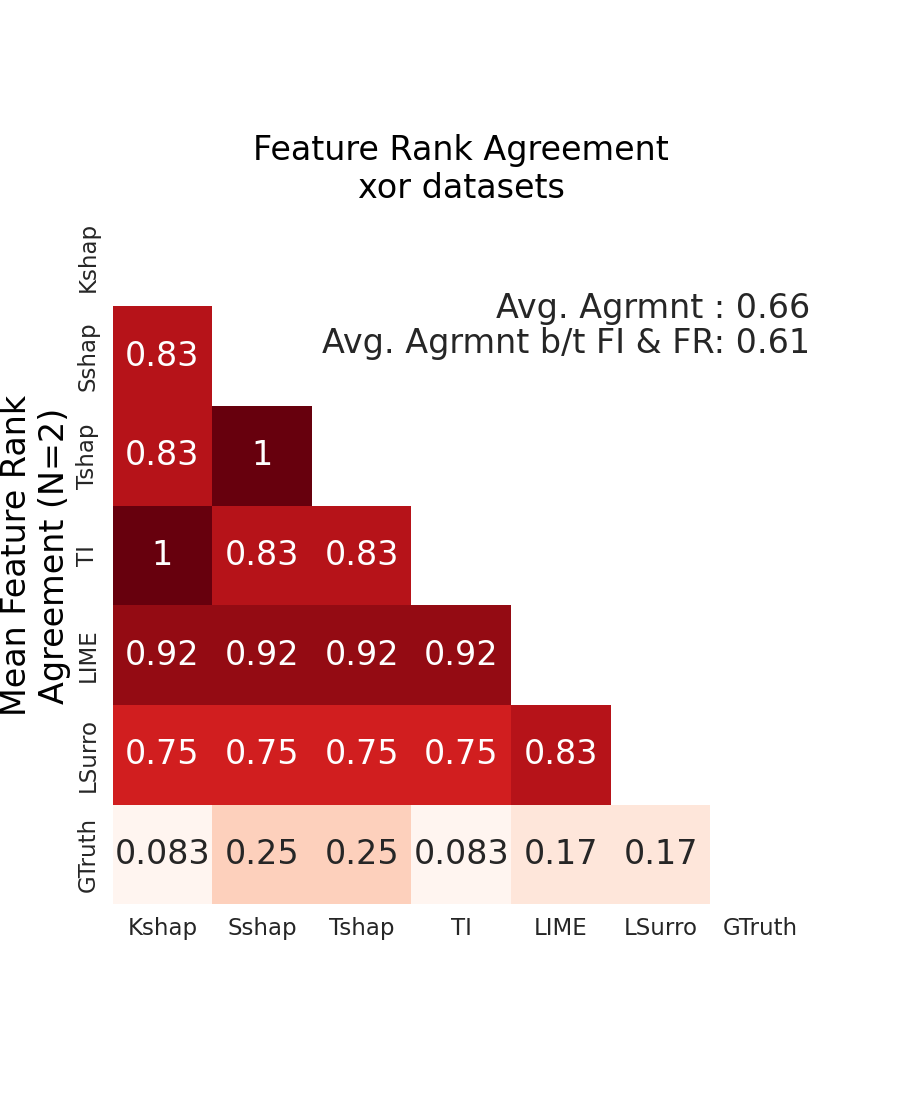}
   \includegraphics[width=\textwidth/3]{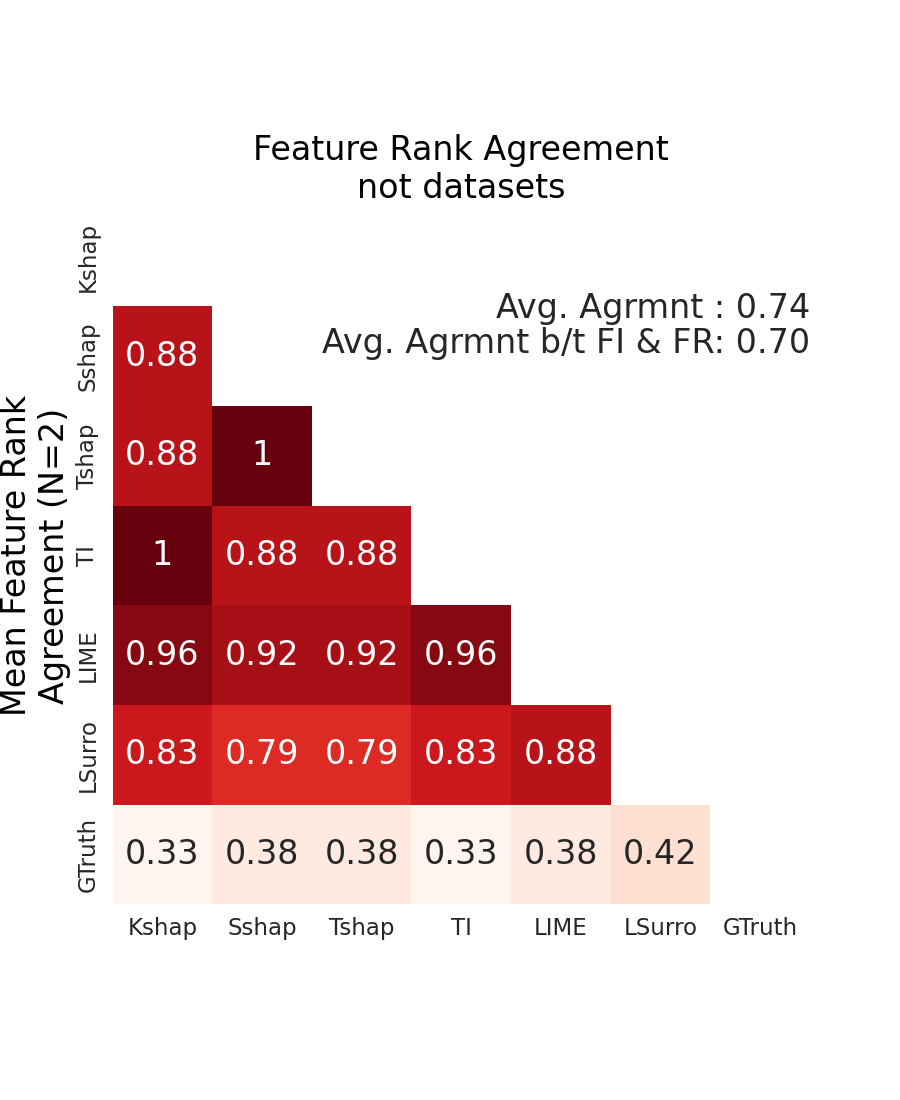}
   \includegraphics[width=\textwidth/3]{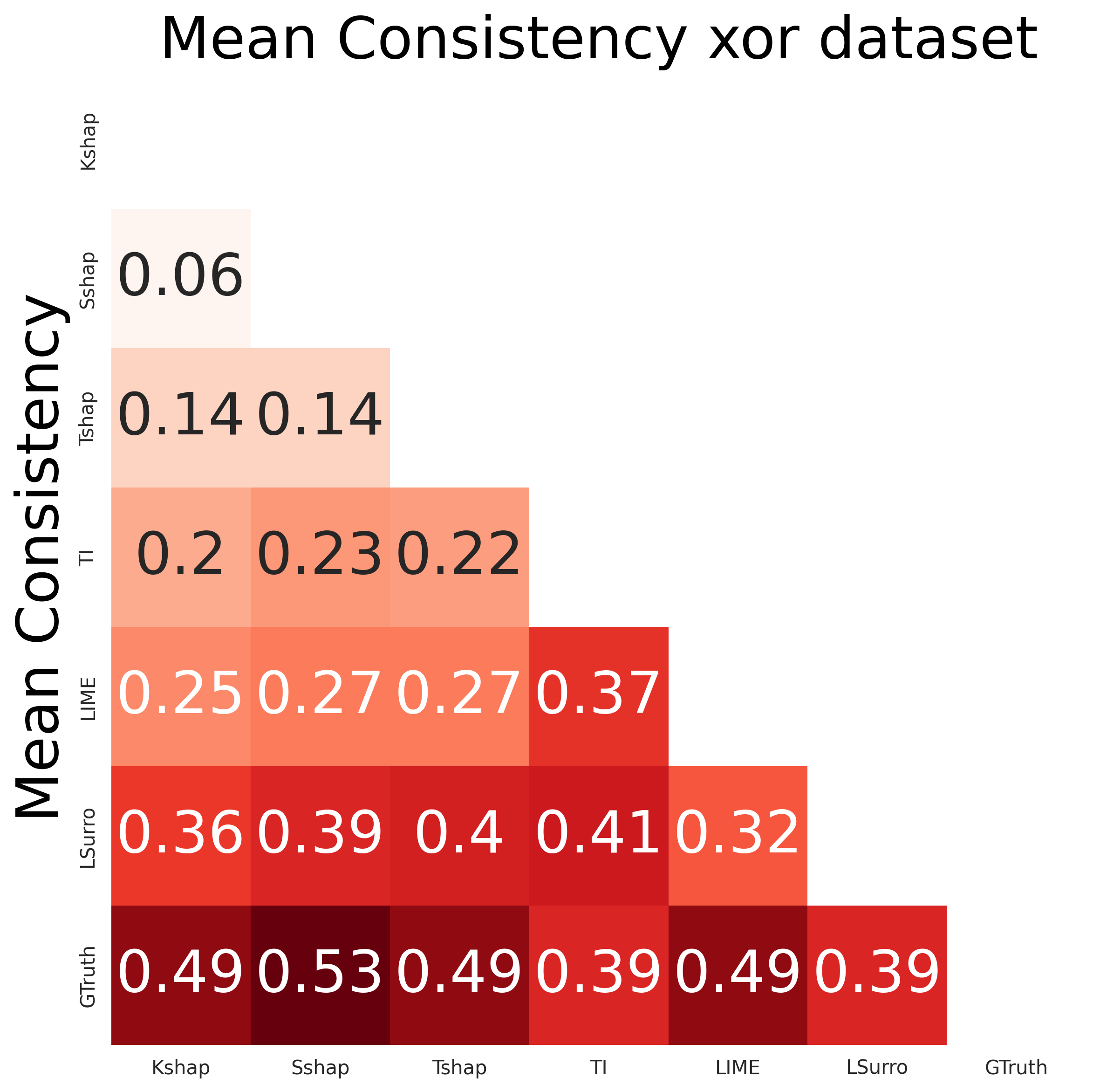}
  \includegraphics[width=\textwidth/3]{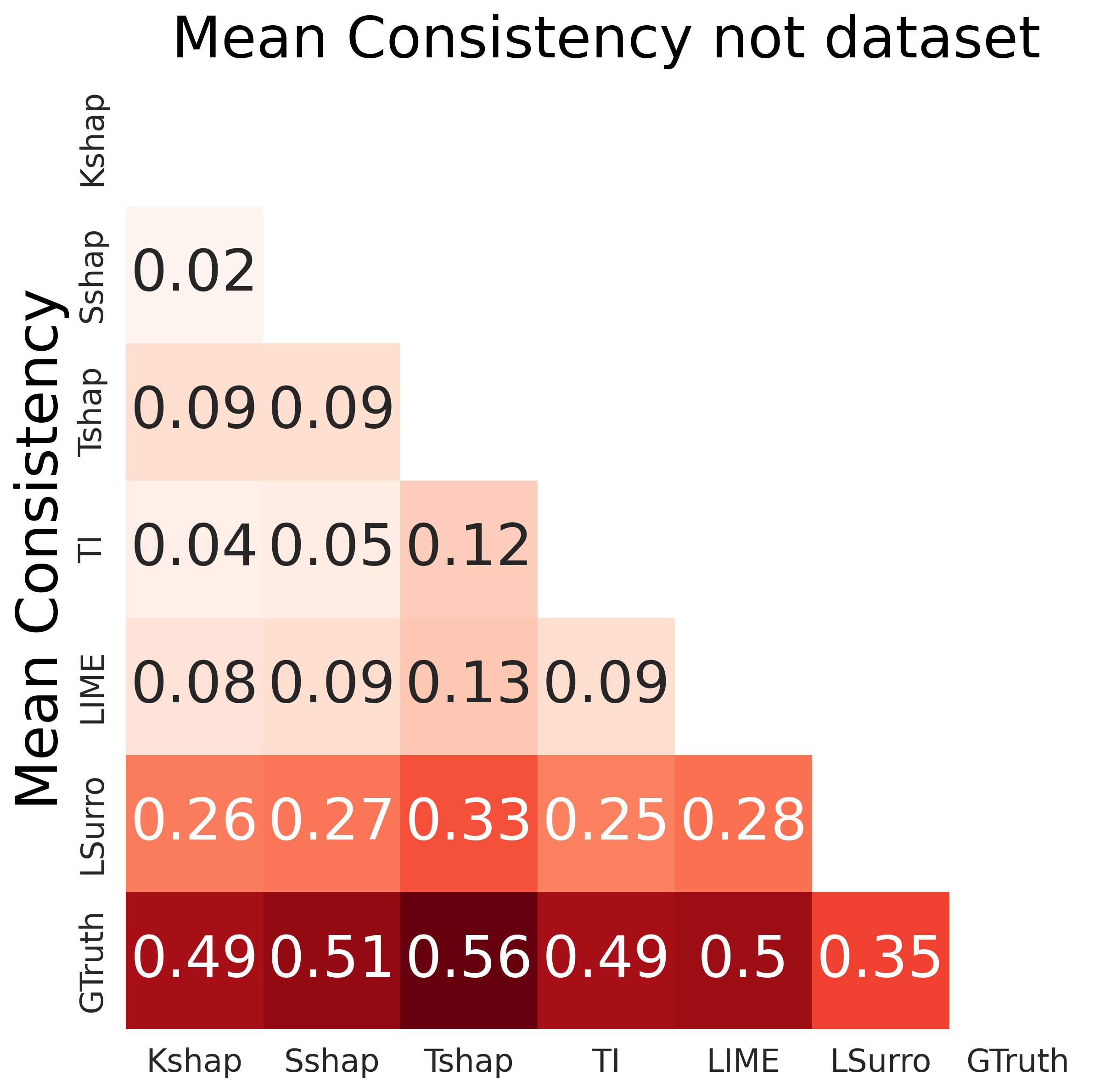}

  \caption{\label{frag}Mean consistency, mean feature agreements for \textsf{XOR} and \textsf{NOT} datasets. Consistency is expressed in $L_2$ distance (the lower the better). Feature agreement measures the fraction of common features between the sets of top-k features of the two rankings (the higher the better). 
  }
\end{figure}

\paragraph{SHAP explainers yield very comparable explanations}
Figure \ref{frag} shows the faithfulness of the explanations to the ground truth measured by mean consistency and mean feature agreements across the \textsf{XOR} and \textsf{NOT} generated datasets.

\textsf{SHAP} explainers yield consistent explanations due to the same feature importance attribution mechanism they all employ. However their explanations are the most inconsistent with respect to the ground truth values. 
Furthermore and for both \textsf{XOR} and \textsf{NOT} datasets, on average the fraction of common feature importance between \textsf{TI} and \textsf{KShap} and between \textsf{SShap} and \textsf{TShap} match perfectly. 


\begin{table}[h]
    \centering
    \resizebox{\columnwidth}{!}{
     \begin{tabular}{|l|c|c|c|c|c|}
    \hline
    function &  Methods  & \# Features for 90\% approximation & Distance with 1 feature(\%) & Mean consistency & Mean Stability (10) \\
    \hline  
    \textsf{XOR} & \textsf{Kshap} &       1.00       &       0.10       &  0.21    &  0.21  \\
    \textsf{XOR} & \textsf{Sshap} &       1.00      &        0.11       &  0.23    &  0.24   \\
    \textsf{XOR} & \textsf{Tshap} &       1.00      &        0.10       &  0.24    &  0.25  \\
    \textsf{XOR} & \textsf{TI} &          1.00      &        0.17       &  0.26    &  0.24   \\
    \textsf{XOR} & \textsf{LIME} &        1.00    &          0.16       &  0.28    &  0.33   \\
    \textsf{XOR} & \textsf{LSurro} &      1.33     &         0.43       &  0.32    &  0.15   \\   
     \hline
    \textsf{NOT} & \textsf{Kshap} &        1.00     &         0.03           &  0.14    &  0.16  \\
    \textsf{NOT}& \textsf{Sshap} &         1.00         &     0.03           &  0.15    &  0.20   \\
    \textsf{NOT}& \textsf{Tshap} &         1.00      &        0.03           &  0.19    &  0.19   \\
    \textsf{NOT}& \textsf{TI} &            1.00    &          0.02           &  0.15    &  0.19   \\
    \textsf{NOT}& \textsf{LIME} &          1.00       &       0.04           &  0.17    &  0.27   \\
    \textsf{NOT}& \textsf{LSurro} &        1.17      &        0.30        &  0.25    &   0.08  \\ 
    \hline
    \end{tabular}}
    \caption{Compactness and stability of the explanations for the \textsf{XOR} datasets. \textsf{Tshap} and \textsf{TI} are the most stable explainers for \textsf{XOR} and \textsf{LSurro} uses only one feature to make nearly half of the prediction of \textsf{NOT}.}
    \label{perfsxor}
\end{table}

   
\paragraph{\textsf{LSurro} is the most locally stable, overestimates unimportant features and achieves better model accuracy with less features}
Table \ref{perfsxor} shows mean consistency, mean stability and compactness across the \textsf{XOR} and \textsf{NOT} datasets.
For \textsf{XOR} and \textsf{NOT} datasets respectively, \textsf{Kshap} is the most consistent to the rest of the explanatory methods on average.
Additionally, on average, \textsf{LSurro} generates the most locally stable explanations in \textsf{XOR} and \textsf{NOT} datasets, achieves higher model estimation and often consider both features as important fr both datasets.

\subsubsection{Real-world data}
For the real-world datasets the ground truth feature importance is unavailable, we perform evaluation of the different metrics in Section \ref{metrics}.

\begin{figure}[h]%
  \centering
  \includegraphics[width=\textwidth/3]{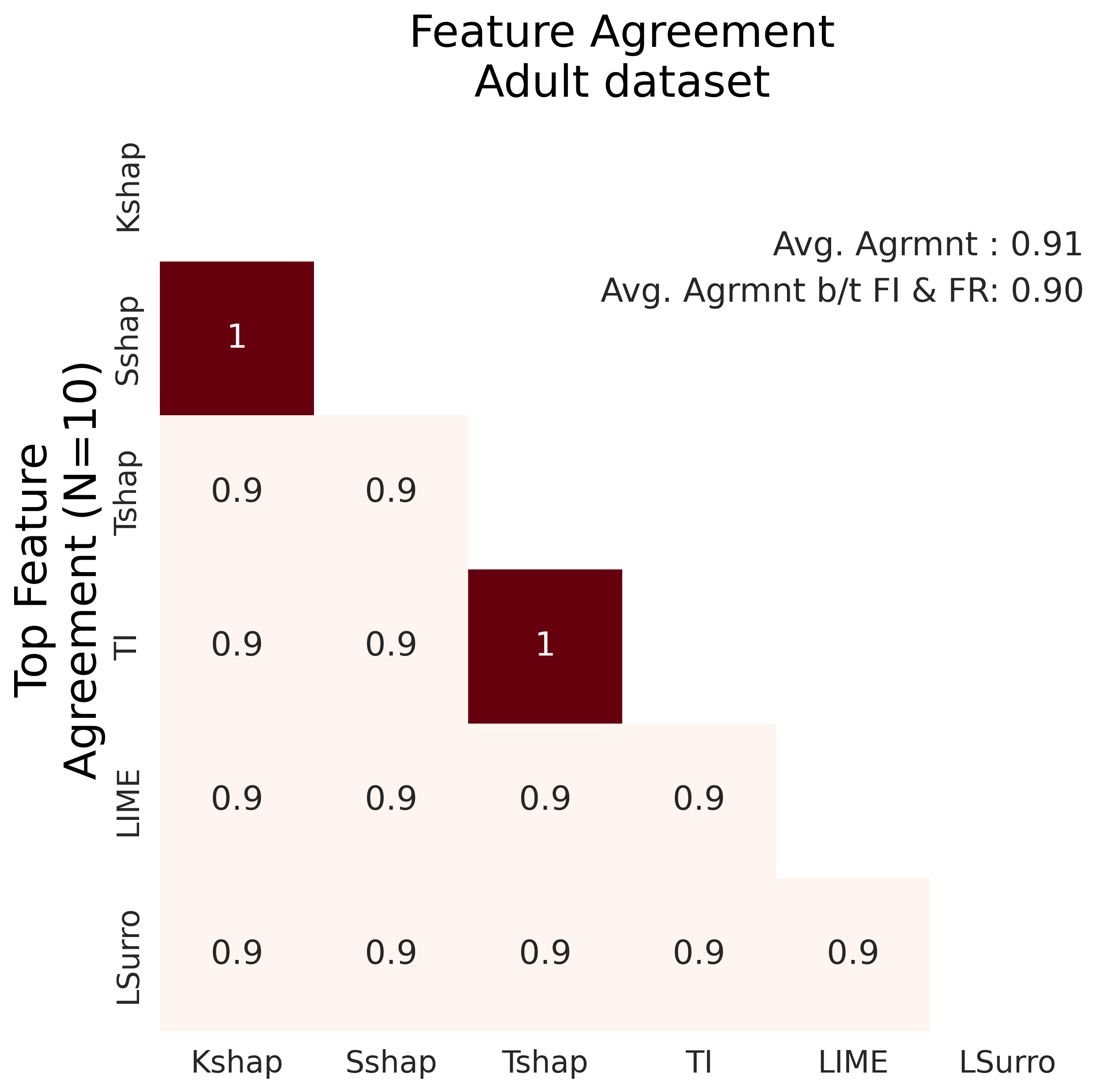}
  \includegraphics[width=\textwidth/3]{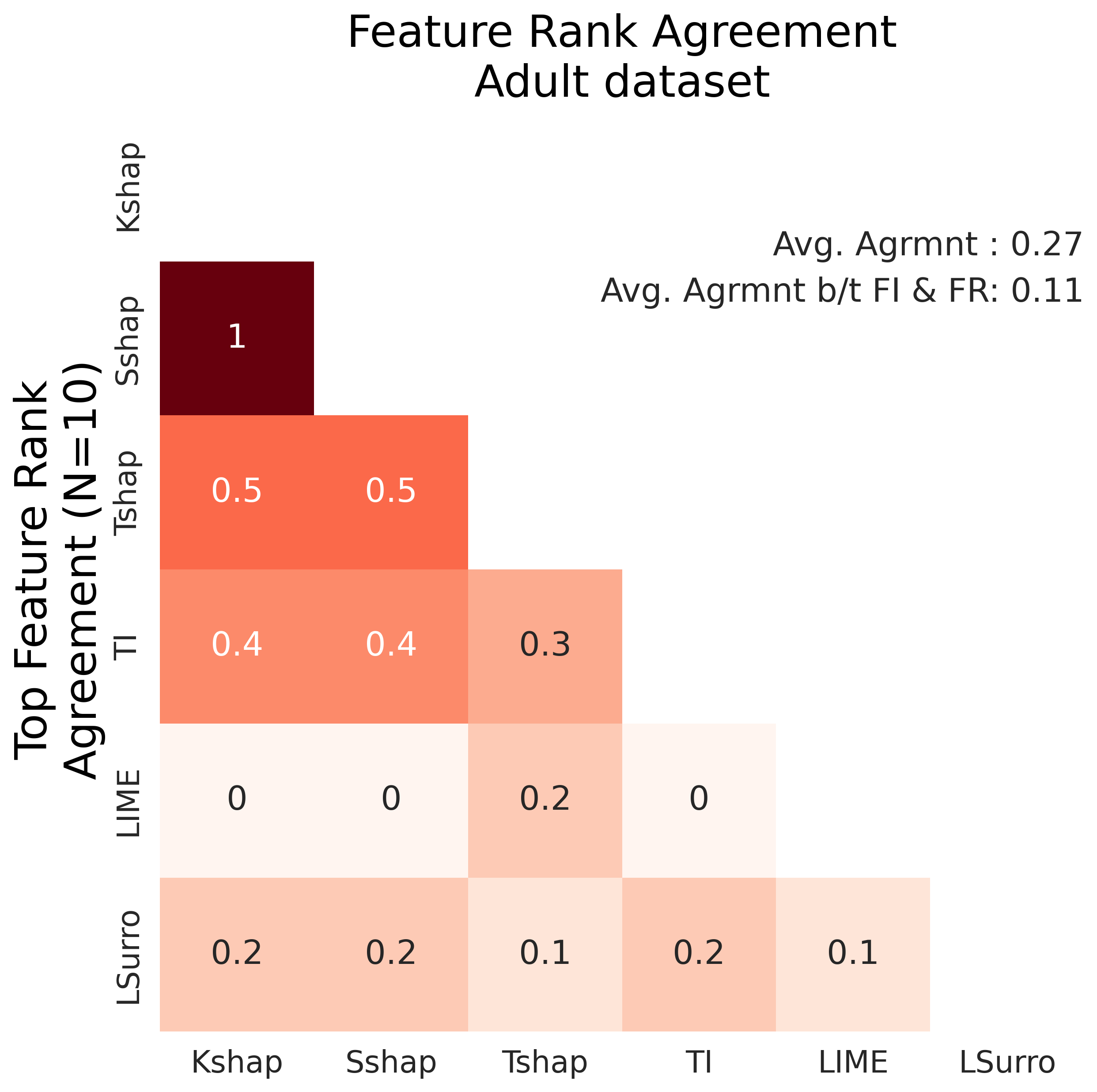}
  \caption{Feature and rank agreements for \textsc{Adult Income} Income dataset. }\label{adult}
\end{figure}

\paragraph{Local surrogates achieves 100\% of model accuracy with 5 features on \textsc{Adult Income} dataset.}
Figure \ref{adult} shows feature agreements for \textsc{Adult Income} Income dataset. 
\textsf{Kshap} and \textsf{Sshap} have exactly the same top-10 feature attributions and ranking. \textsf{TI} and \textsf{Tshap} share the same set of top-10 features. The rest of the explainers share 90\% of the top 10 most important features.
\textsf{LIME} share the lowest of top-10 important features with TI, \textsf{Kshap} and \textsf{Sshap} on the ranking of the top 10 features.
Additionally, Table \ref{adult2} illustrates the compactness, mean consistency and stability of the different methods on the \textsc{Adult Income} Income dataset. \textsf{LSurro} explains 100\% of model output with only 5 features. \textsf{Kshap}, \textsf{Tshap} and \textsf{TI} are the most stable for this dataset and \textsf{LIME} have the highest mean consistency across all the datasets.

\begin{table}[h]
    \centering
    \resizebox{\columnwidth}{!}{ 
    \begin{tabular}{|l|c|c|c|c|}
    \hline
    Methods  & \# Features for 90\% Accuracy & Accuracy with 5 feature(\%) & Mean consistency & Mean Stability \\
    \hline
    \textsf{Kshap} &          \textbf{1} &          09 &  0.43 & \textbf{0.00}\\
    \textsf{Tshap} &         \textbf{1} &          09 & 0.43  & \textbf{0.00} \\
    \textsf{Sshap} &        \textbf{1}&          08 &  0.43 & 0.01 \\
    \textsf{LIME} &         \textbf{1} &          07 & \textbf{0.15}  & 0.07 \\
    \textsf{TI} &          \textbf{1}&          10 &  0.5    &  \textbf{0.00} \\
    \textsf{LSurro} &              3 &      \textbf{100} & 0.7 &  0.01 \\
    \hline
    \end{tabular}
    }
    \caption{Compactness, mean consistency and stability for \textsc{Adult Income} dataset.}
    \label{adult2}
\end{table}

\begin{figure}[h]
  \centering
  \includegraphics[width=\textwidth/3]{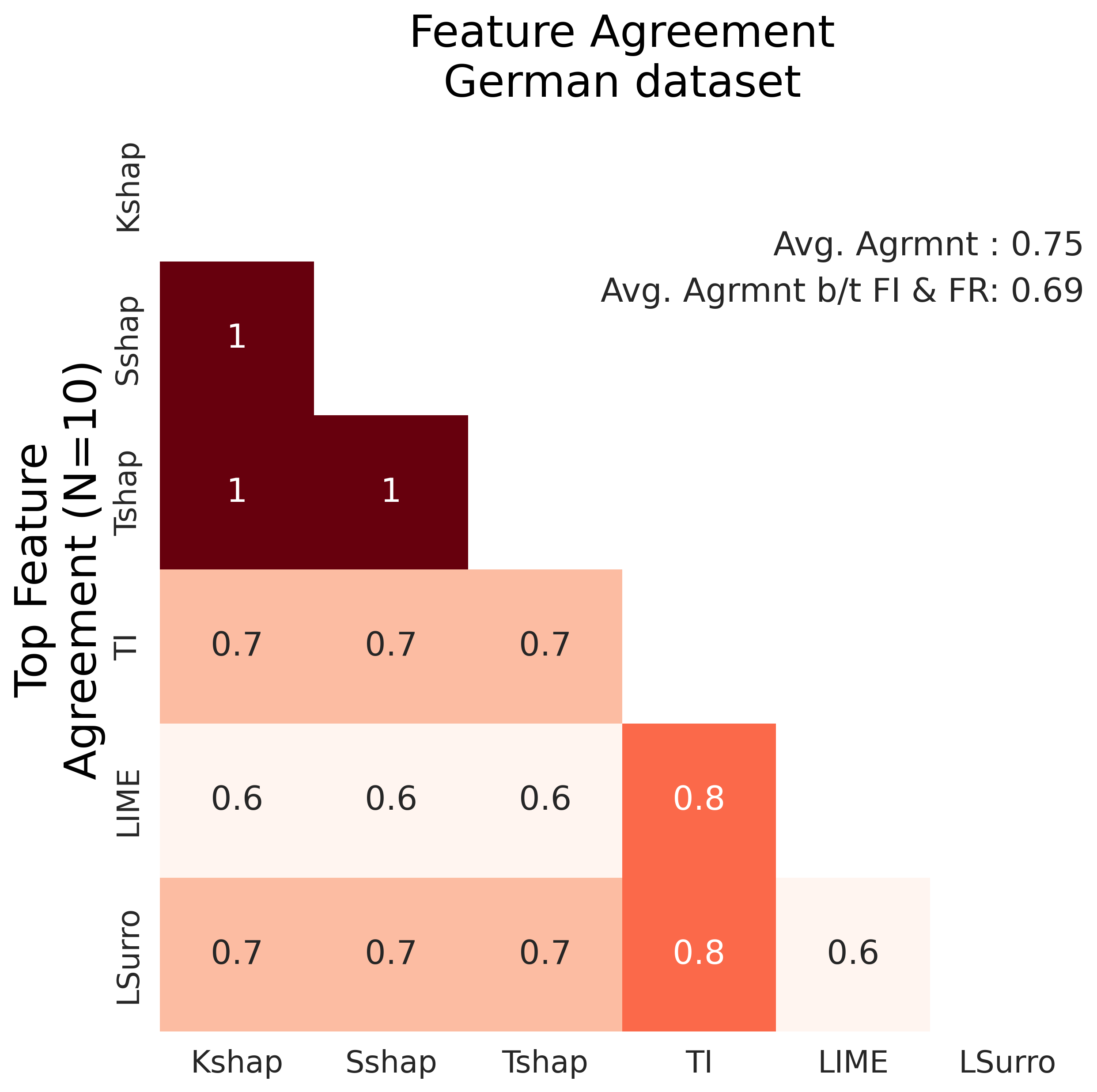}
 \includegraphics[width=\textwidth/3]{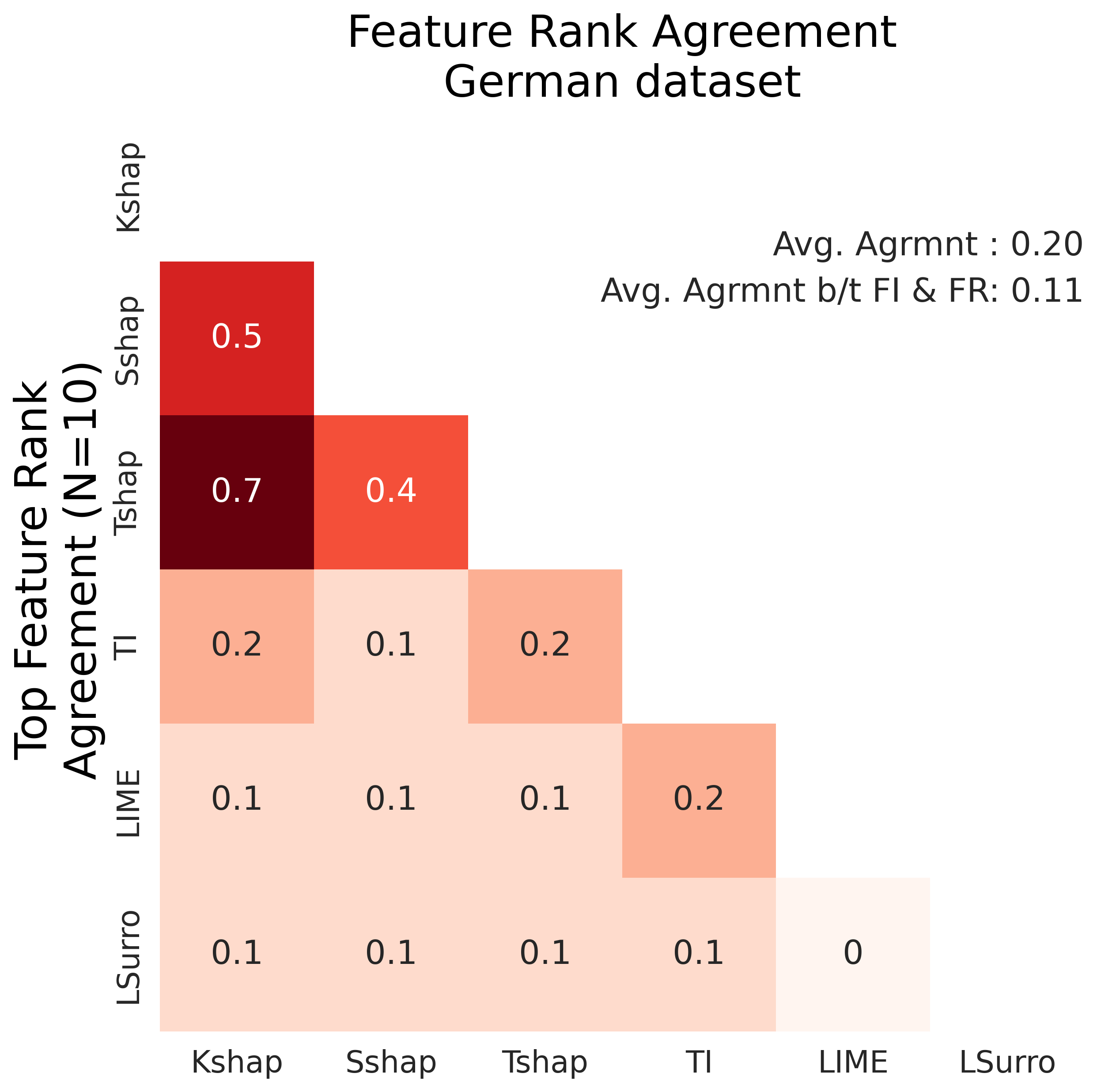}
  \caption{\label{german} Feature and rank agreements for \textsc{German Credit Risk} dataset. }
\end{figure}

\paragraph{SHAP explainers generate the most consistent explanations for \textsc{German Credit Risk} dataset.}
Figure \ref{german} and Table \ref{german2} show the computed metrics on \textsc{German Credit Risk} dataset. Overall, all the explainers achieve 90\% of model accuracy with only 1 feature and \textsf{SHAP} explainers are the most consistent among the methods. \textsf{LIME} is the most stable among the explainers. Moreover, \textsf{SHAP} explainers have same top-10 most important features as they all use the same mechanism of computation of the Shapley values the feature importance estimates, \textsf{LIME} have only 6 top features in common with  \textsf{SHAP} explainers, and \textsf{Kshap} and \textsf{Tshap} share 7 features with the same rankings.

\begin{table}[h]
    \centering
    \resizebox{\columnwidth}{!}{ 
    \begin{tabular}{|l|c|c|c|c|}
    \hline
    Methods  & \# Features for 90\% Accuracy & Accuracy with 5 feature(\%) & Mean consistency & Mean Stability \\
    \hline
    
    \textsf{Kshap} &           \textbf{1} &          12 &  \textbf{0.45} & 0.03\\
    \textsf{Tshap} &           \textbf{1}&          13 &  \textbf{0.45}  & 0.03\\
    \textsf{Sshap} &           \textbf{1} &          12 &  \textbf{0.45}  & 0.03\\
    \textsf{LIME} &          \textbf{1}&          41 &  1.18  &  \textbf{0.00}\\
    \textsf{TI} &           \textbf{1} &          13 &   0.54   & 0.02\\
    \textsf{LSurro} &        \textbf{1} &      \textbf{57} & 0.74 & 0.03 \\
    \hline
    \end{tabular}
    }
    \caption{Compactness, mean consistency and stability for \textsc{German Credit Risk} dataset.}
    \label{german2}
\end{table}

\begin{figure}[h]
  \centering
  \includegraphics[width=\textwidth/3]{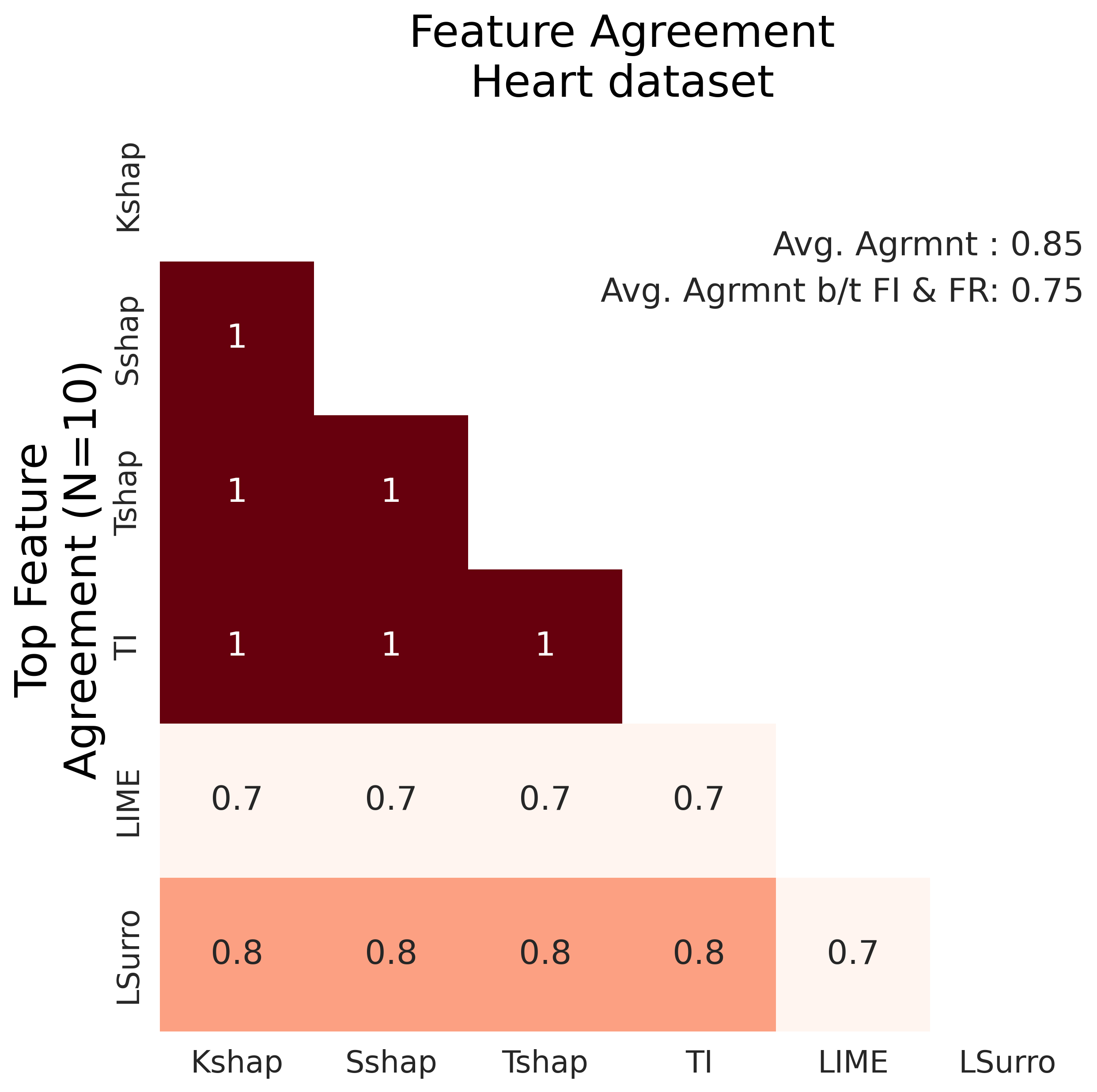}
  \includegraphics[width=\textwidth/3]{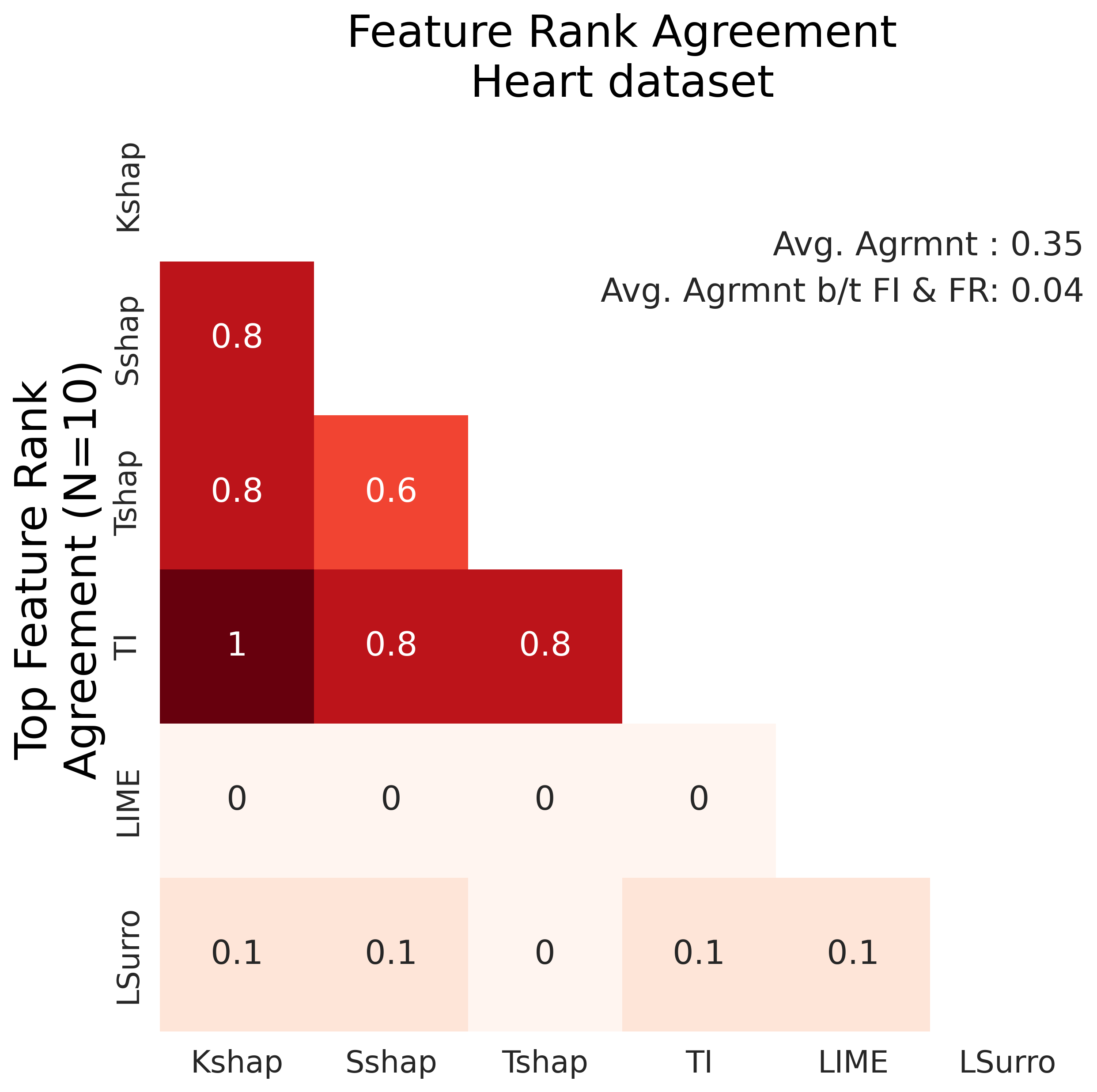}
  \caption{\label{heart}Feature and rank agreements for \textsc{Heart Diagnosis} dataset.}
\end{figure}

\paragraph{SHAP explainers and \textsf{TI} share the top-10 most important feature for \textsc{Heart Diagnosis} dataset}
In Figure \ref{heart} and Table \ref{heart2}, \textsf{SHAP} explainers and \textsf{TI} share the top-10 most important feature, contrary to \textsf{LIME} which doesn't share same ranking of the top features with other explainers. \textsf{Kshap} and \textsf{TI} have exactly the same top-10 features and in the same rankings.
On the other hand, \textsf{LSurro} estimates 100\% of model accuracy with only 5 features and \textsf{SHAP} explainers and \textsf{TI} generate the most stable and consistent explanations.

\begin{table}[h]
    \centering
    \resizebox{\columnwidth}{!}{ 
    \begin{tabular}{|l|c|c|c|c|}   
    \hline
   Methods  & \# Features for 90\% Accuracy & Accuracy with 5 feature(\%) & Mean consistency & Mean Stability \\
    \hline
    \textsf{Kshap} &            \textbf{1} &          18 & \textbf{0.4}   & \textbf{0.00} \\
    \textsf{Tshap} &             \textbf{1} &          21 & \textbf{0.4}   & \textbf{0.00} \\
    \textsf{Sshap} &            \textbf{1} &          18 & \textbf{0.4}   & \textbf{0.00} \\
    \textsf{LIME} &            \textbf{1} &          31 &  1.2   &  0.03 \\
    \textsf{TI} &            \textbf{1} &          25 &    0.45   & \textbf{0.00}\\
    \textsf{LSurro}&              6 &       \textbf{100} & 0.63  & 0.03 \\
    \hline
    \end{tabular}
    }
    \caption{Compactness, mean consistency and stability for \textsc{Heart Diagnosis} dataset.}
    \label{heart2}
\end{table}

\begin{figure}[h]
  \centering
  \includegraphics[width=\textwidth/3]{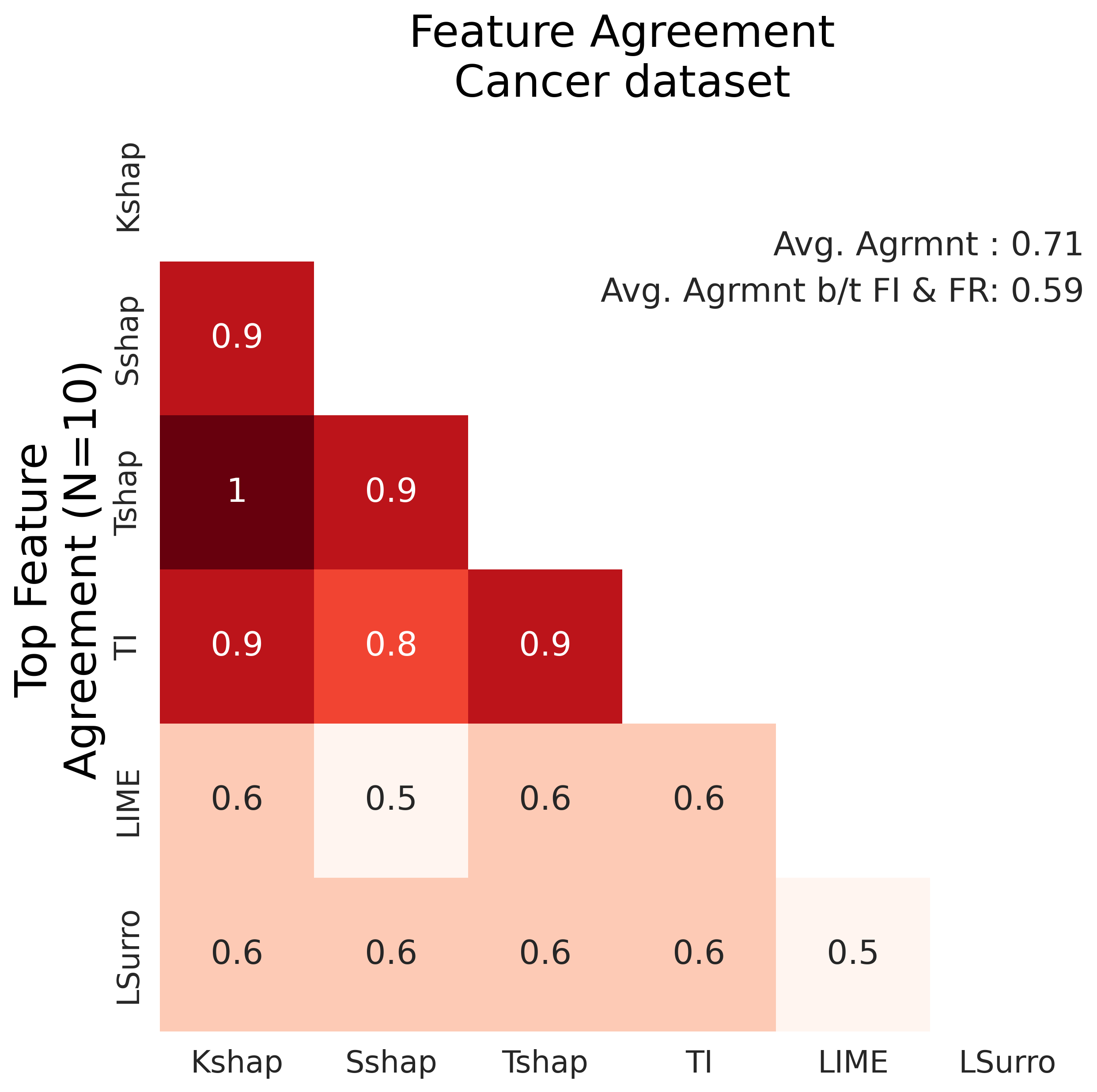}
  \includegraphics[width=\textwidth/3]{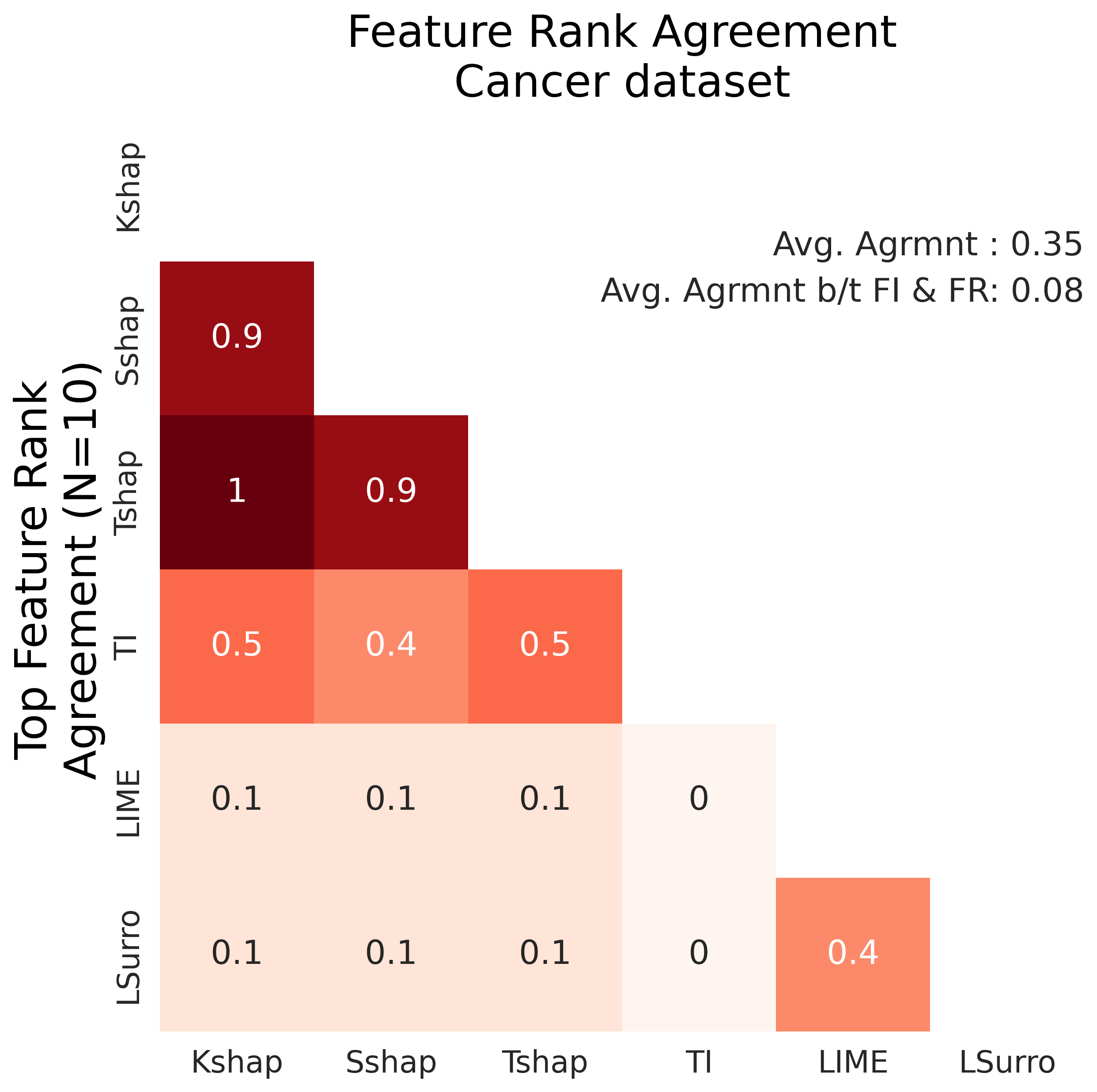}
  \caption{\label{cancer}Feature and rank agreements for \textsc{Cervical Cancer} dataset.}
\end{figure}

\paragraph{Local surrogates are the most consistent and explains 100\% of model output with 5 features for \textsc{Cervical Cancer} dataset.}
Figure \ref{cancer} and Table \ref{cancer2} illustrate above metrics on the \textsc{Cervical Cancer} dataset.
\textsf{Tshap} and \textsf{Kshap} share the top 10 features. \textsf{LIME} and \textsf{Sshap} have the lowest top features in common. \textsf{LIME} and \textsf{LSurro} have no comparable rankings of the the features with TI, although they both learn a surrogate linear model in the neighborhood of each instance but use different mechanisms of the generation of the local neighborhood, which can explain the disagreement in their features importance estimates.
Additionally, \textsf{LSurro} is the most consistent and explains 100\% of model output with 5 features, and \textsf{SHAP} explainers have the highest mean stability.

\begin{table}[h]
    \centering
    \resizebox{\columnwidth}{!}{  
    \begin{tabular}{|l|c|c|c|c|}
    \hline
       Methods  & \# Features for 90\% Accuracy & Accuracy with 5 feature(\%) & Mean consistency & Mean Stability \\
        \hline
         \textsf{Kshap} & \textbf{1}& 12  & 1.44 & \textbf{0.34}\\
         \textsf{Sshap}  & \textbf{1}& 12 & 0.87 & \textbf{0.34}\\
         \textsf{Tshap}  & \textbf{1}& 12  & 0.36  & \textbf{0.34}\\
         \textsf{LIME}   & \textbf{1}& 45 & 0.66 & 0.46\\
         \textsf{TI}    & \textbf{1}&  19  & 0.59 &  0.42\\
         \textsf{LSurro}  & 3&  \textbf{100}  & \textbf{0.22} & 0.54\\
         \hline
    \end{tabular}
    }
    \caption{Compactness, stability and consistency of local explainability methods for predicting \textsc{Cervical Cancer} risk. Some methods predominantly require only one feature to achieve 90\% prediction accuracy. \textsf{LSurro} have the highest mean stability, while  \textsf{SHAP} variants have the highest mean consistency.}
    \label{cancer2}
\end{table}



\section{Conclusions and Future Work}
The experiments on the synthetic and real-world datasets showed that feature importance attribution can be affected by multiple factors such as data properties, the black-box model and the assumptions on which the explainable method is built to attribute feature contributions. We restricted our study to the first factor with a focus on some data properties, decision tree based models, on tabular data and for a binary classification task.  
For the matter of simplicity and ease of understanding of the model's behavior, we restricted our generation model to two variables in order to easily track the feature interactions, which can be a limit in real-world scenarios because of the need to handle high-dimensional data in many situations.  
\begin{itemize}
    \item For datasets with irrelevant variables, avoid using \textsf{LSurro} and \textsf{LIME} because both overestimate the importance of unimportant features.
    \item We recommend to avoid using \textsf{TI} for highly noisy datasets because \textsf{TI} is the most unstable compared to other explainers in such datasets. This can be justified by the decomposition of of the feature importance used by \textsf{TI}, which allocates importance to the noise.
    \item Kernel, Sampling and Tree  \textsf{SHAP} explainers give very similar explanations, thus we recommend using \textsf{Sshap} or \textsf{Tshap} for faster computations and adaptability for decision trees. 
\end{itemize}

\paragraph{Perspectives} Future work should further focus on each single explainability method separately to be able to explore in depth the effect of its assumptions and its inner workings for a single model parameter and on one specific data property on the feature importance attribution. Also, it is of interest to assess these feature importance estimates on other data parameters such as the number of instances in the test set and the number of features.
Our work can be applied on other tasks such as regression and multi-class classification, to image and text data.

\newpage
\bibliography{mlj}


\begin{thebibliography}{37}
\ifx \bisbn   \undefined \def \bisbn  #1{ISBN #1}\fi
\ifx \binits  \undefined \def \binits#1{#1}\fi
\ifx \bauthor  \undefined \def \bauthor#1{#1}\fi
\ifx \batitle  \undefined \def \batitle#1{#1}\fi
\ifx \bjtitle  \undefined \def \bjtitle#1{#1}\fi
\ifx \bvolume  \undefined \def \bvolume#1{\textbf{#1}}\fi
\ifx \byear  \undefined \def \byear#1{#1}\fi
\ifx \bissue  \undefined \def \bissue#1{#1}\fi
\ifx \bfpage  \undefined \def \bfpage#1{#1}\fi
\ifx \blpage  \undefined \def \blpage #1{#1}\fi
\ifx \burl  \undefined \def \burl#1{\textsf{#1}}\fi
\ifx \doiurl  \undefined \def \doiurl#1{\url{https://doi.org/#1}}\fi
\ifx \betal  \undefined \def \betal{\textit{et al.}}\fi
\ifx \binstitute  \undefined \def \binstitute#1{#1}\fi
\ifx \binstitutionaled  \undefined \def \binstitutionaled#1{#1}\fi
\ifx \bctitle  \undefined \def \bctitle#1{#1}\fi
\ifx \beditor  \undefined \def \beditor#1{#1}\fi
\ifx \bpublisher  \undefined \def \bpublisher#1{#1}\fi
\ifx \bbtitle  \undefined \def \bbtitle#1{#1}\fi
\ifx \bedition  \undefined \def \bedition#1{#1}\fi
\ifx \bseriesno  \undefined \def \bseriesno#1{#1}\fi
\ifx \blocation  \undefined \def \blocation#1{#1}\fi
\ifx \bsertitle  \undefined \def \bsertitle#1{#1}\fi
\ifx \bsnm \undefined \def \bsnm#1{#1}\fi
\ifx \bsuffix \undefined \def \bsuffix#1{#1}\fi
\ifx \bparticle \undefined \def \bparticle#1{#1}\fi
\ifx \barticle \undefined \def \barticle#1{#1}\fi
\bibcommenthead
\ifx \bconfdate \undefined \def \bconfdate #1{#1}\fi
\ifx \botherref \undefined \def \botherref #1{#1}\fi
\ifx \url \undefined \def \url#1{\textsf{#1}}\fi
\ifx \bchapter \undefined \def \bchapter#1{#1}\fi
\ifx \bbook \undefined \def \bbook#1{#1}\fi
\ifx \bcomment \undefined \def \bcomment#1{#1}\fi
\ifx \oauthor \undefined \def \oauthor#1{#1}\fi
\ifx \citeauthoryear \undefined \def \citeauthoryear#1{#1}\fi
\ifx \endbibitem  \undefined \def \endbibitem {}\fi
\ifx \bconflocation  \undefined \def \bconflocation#1{#1}\fi
\ifx \arxivurl  \undefined \def \arxivurl#1{\textsf{#1}}\fi
\csname PreBibitemsHook\endcsname

\bibitem[\protect\citeauthoryear{Breiman}{2001}]{breimanRandomForests2001a}
\begin{barticle}
\bauthor{\bsnm{Breiman}, \binits{L.}}:
\batitle{Random {{Forests}}}.
\bjtitle{Machine Learning}
\bvolume{45}(\bissue{1}),
\bfpage{5}--\blpage{32}
(\byear{2001})
\end{barticle}
\endbibitem

\bibitem[\protect\citeauthoryear{Kulmala et~al.}{2017}]{Multitree}
\begin{barticle}
\bauthor{\bsnm{Kulmala}, \binits{L.}},
\bauthor{\bsnm{Read}, \binits{J.}},
\bauthor{\bsnm{Nöjd}, \binits{P.}},
\bauthor{\bsnm{Rathgeber}, \binits{C.B.K.}},
\bauthor{\bsnm{Cuny}, \binits{H.E.}},
\bauthor{\bsnm{Hollmén}, \binits{J.}},
\bauthor{\bsnm{Mäkinen}, \binits{H.}}:
\batitle{Identifying the main drivers for the production and maturation of
  scots pine tracheids along a temperature gradient}.
\bjtitle{Agricultural and Forest Meteorology}
\bvolume{232}(\bissue{January}),
\bfpage{210}--\blpage{224}
(\byear{2017})
\end{barticle}
\endbibitem

\bibitem[\protect\citeauthoryear{Ricordeau and
  Lacaille}{2010}]{ricordeauApplicationRandomForests2010}
\begin{bchapter}
\bauthor{\bsnm{Ricordeau}, \binits{J.}},
\bauthor{\bsnm{Lacaille}, \binits{J.}}:
\bctitle{Application of {{Random Forests}} to {{Engine}} health
  {{Monitoring}}}.
(\byear{2010})
\end{bchapter}
\endbibitem

\bibitem[\protect\citeauthoryear{Strobl
  et~al.}{2007}]{stroblBiasRandomForest2007}
\begin{barticle}
\bauthor{\bsnm{Strobl}, \binits{C.}},
\bauthor{\bsnm{Boulesteix}, \binits{A.-L.}},
\bauthor{\bsnm{Zeileis}, \binits{A.}},
\bauthor{\bsnm{Hothorn}, \binits{T.}}:
\batitle{Bias in random forest variable importance measures: {{Illustrations}},
  sources and a solution}.
\bjtitle{BMC Bioinformatics}
\bvolume{8}(\bissue{1}),
\bfpage{25}
(\byear{2007})
\end{barticle}
\endbibitem

\bibitem[\protect\citeauthoryear{Terence and Prince}{}]{BewareDefaultRandom}
\begin{botherref}
\oauthor{\bsnm{Terence}, \binits{P.}},
\oauthor{\bsnm{Prince}, \binits{G.}}:
Beware {{Default Random Forest Importances}}.
http://explained.ai/decision-tree-viz/index.html
\end{botherref}
\endbibitem

\bibitem[\protect\citeauthoryear{Molnar
  et~al.}{2020}]{LimitationsInterpretableMachine}
\begin{bbook}
\bauthor{\bsnm{Molnar}, \binits{C.}},
\bauthor{\bsnm{Gruber}, \binits{S.}},
\bauthor{\bsnm{Kopper}, \binits{P.}}:
\bbtitle{Limitations of {{Interpretable Machine Learning Methods}}},
(\byear{2020})
\end{bbook}
\endbibitem

\bibitem[\protect\citeauthoryear{Ribeiro
  et~al.}{2016}]{ribeiroWhyShouldTrust2016}
\begin{bchapter}
\bauthor{\bsnm{Ribeiro}, \binits{M.T.}},
\bauthor{\bsnm{Singh}, \binits{S.}},
\bauthor{\bsnm{Guestrin}, \binits{C.}}:
\bctitle{"{{Why Should I Trust You}}?": {{Explaining}} the {{Predictions}} of
  {{Any Classifier}}}.
In: \bbtitle{Proceedings of the 22nd {{ACM SIGKDD International Conference}} on
  {{Knowledge Discovery}} and {{Data Mining}}}.
\bsertitle{{{KDD}} '16},
pp. \bfpage{1135}--\blpage{1144}.
\bpublisher{{Association for Computing Machinery}},
\blocation{{New York, NY, USA}}
(\byear{2016})
\end{bchapter}
\endbibitem

\bibitem[\protect\citeauthoryear{Lundberg and
  Lee}{2017}]{lundbergUnifiedApproachInterpreting2017}
\begin{botherref}
\oauthor{\bsnm{Lundberg}, \binits{S.}},
\oauthor{\bsnm{Lee}, \binits{S.-I.}}:
A {{Unified Approach}} to {{Interpreting Model Predictions}}.
arXiv:1705.07874 [cs, stat]
(2017)
{\href{https://arxiv.org/abs/1705.07874}{{arXiv:1705.07874}}}
{[cs, stat]}
\end{botherref}
\endbibitem

\bibitem[\protect\citeauthoryear{Krishna
  et~al.}{2022}]{krishnaDisagreementProblemExplainable2022}
\begin{botherref}
\oauthor{\bsnm{Krishna}, \binits{S.}},
\oauthor{\bsnm{Han}, \binits{T.}},
\oauthor{\bsnm{Gu}, \binits{A.}},
\oauthor{\bsnm{Pombra}, \binits{J.}},
\oauthor{\bsnm{Jabbari}, \binits{S.}},
\oauthor{\bsnm{Wu}, \binits{S.}},
\oauthor{\bsnm{Lakkaraju}, \binits{H.}}:
The {{Disagreement Problem}} in {{Explainable Machine Learning}}: {{A
  Practitioner}}'s {{Perspective}}.
{arXiv}
(2022)
\end{botherref}
\endbibitem

\bibitem[\protect\citeauthoryear{Attanasio
  et~al.}{2022}]{attanasioFerretFrameworkBenchmarking2022}
\begin{botherref}
\oauthor{\bsnm{Attanasio}, \binits{G.}},
\oauthor{\bsnm{Pastor}, \binits{E.}},
\oauthor{\bsnm{Di~Bonaventura}, \binits{C.}},
\oauthor{\bsnm{Nozza}, \binits{D.}}:
Ferret: A {{Framework}} for {{Benchmarking Explainers}} on {{Transformers}}.
{arXiv}
(2022)
\end{botherref}
\endbibitem

\bibitem[\protect\citeauthoryear{Camburu
  et~al.}{2019}]{camburuCanTrustExplainer2019}
\begin{botherref}
\oauthor{\bsnm{Camburu}, \binits{O.-M.}},
\oauthor{\bsnm{Giunchiglia}, \binits{E.}},
\oauthor{\bsnm{Foerster}, \binits{J.}},
\oauthor{\bsnm{Lukasiewicz}, \binits{T.}},
\oauthor{\bsnm{Blunsom}, \binits{P.}}:
Can {{I Trust}} the {{Explainer}}? {{Verifying Post-hoc Explanatory Methods}}.
{arXiv}
(2019)
\end{botherref}
\endbibitem

\bibitem[\protect\citeauthoryear{Bodria
  et~al.}{2021}]{bodriaBenchmarkingSurveyExplanation2021}
\begin{botherref}
\oauthor{\bsnm{Bodria}, \binits{F.}},
\oauthor{\bsnm{Giannotti}, \binits{F.}},
\oauthor{\bsnm{Guidotti}, \binits{R.}},
\oauthor{\bsnm{Naretto}, \binits{F.}},
\oauthor{\bsnm{Pedreschi}, \binits{D.}},
\oauthor{\bsnm{Rinzivillo}, \binits{S.}}:
Benchmarking and {{Survey}} of {{Explanation Methods}} for {{Black Box
  Models}}.
{arXiv}
(2021)
\end{botherref}
\endbibitem

\bibitem[\protect\citeauthoryear{Neely
  et~al.}{2021}]{neelyOrderCourtExplainable2021}
\begin{botherref}
\oauthor{\bsnm{Neely}, \binits{M.}},
\oauthor{\bsnm{Schouten}, \binits{S.F.}},
\oauthor{\bsnm{Bleeker}, \binits{M.J.R.}},
\oauthor{\bsnm{Lucic}, \binits{A.}}:
Order in the {{Court}}: {{Explainable AI Methods Prone}} to {{Disagreement}}.
{arXiv}
(2021)
\end{botherref}
\endbibitem

\bibitem[\protect\citeauthoryear{Flora et~al.}{2022}]{flora2022comparing}
\begin{botherref}
\oauthor{\bsnm{Flora}, \binits{M.}},
\oauthor{\bsnm{Potvin}, \binits{C.}},
\oauthor{\bsnm{McGovern}, \binits{A.}},
\oauthor{\bsnm{Handler}, \binits{S.}}:
Comparing explanation methods for traditional machine learning models part 1:
  An overview of current methods and quantifying their disagreement.
arXiv preprint arXiv:2211.08943
(2022)
\end{botherref}
\endbibitem

\bibitem[\protect\citeauthoryear{Molnar}{2022}]{molnar2022}
\begin{bbook}
\bauthor{\bsnm{Molnar}, \binits{C.}}:
\bbtitle{Interpretable Machine Learning},
\bedition{2}nd edn.
(\byear{2022}).
\burl{https://christophm.github.io/interpretable-ml-book}
\end{bbook}
\endbibitem

\bibitem[\protect\citeauthoryear{Lundberg
  et~al.}{2019}]{lundbergConsistentIndividualizedFeature2019a}
\begin{botherref}
\oauthor{\bsnm{Lundberg}, \binits{S.M.}},
\oauthor{\bsnm{Erion}, \binits{G.G.}},
\oauthor{\bsnm{Lee}, \binits{S.-I.}}:
Consistent {{Individualized Feature Attribution}} for {{Tree Ensembles}}.
{arXiv}
(2019)
\end{botherref}
\endbibitem

\bibitem[\protect\citeauthoryear{Li et~al.}{2019}]{treeinterpreter}
\begin{botherref}
\oauthor{\bsnm{Li}, \binits{X.}},
\oauthor{\bsnm{Wang}, \binits{Y.}},
\oauthor{\bsnm{Basu}, \binits{S.}},
\oauthor{\bsnm{Kumbier}, \binits{K.}},
\oauthor{\bsnm{Yu}, \binits{B.}}:
A {Debiased} {MDI} {Feature} {Importance} {Measure} for {Random} {Forests}.
arXiv:1906.10845 [cs, stat]
(2019).
arXiv: 1906.10845.
Accessed 2019-10-18
\end{botherref}
\endbibitem

\bibitem[\protect\citeauthoryear{{Alvarez-Melis} and
  Jaakkola}{2018}]{alvarez-melisRobustnessInterpretabilityMethods2018}
\begin{botherref}
\oauthor{\bsnm{{Alvarez-Melis}}, \binits{D.}},
\oauthor{\bsnm{Jaakkola}, \binits{T.S.}}:
On the {{Robustness}} of {{Interpretability Methods}}.
{arXiv}
(2018)
\end{botherref}
\endbibitem

\bibitem[\protect\citeauthoryear{Rajbahadur
  et~al.}{2022}]{rajbahadurImpactFeatureImportance2022}
\begin{barticle}
\bauthor{\bsnm{Rajbahadur}, \binits{G.K.}},
\bauthor{\bsnm{Wang}, \binits{S.}},
\bauthor{\bsnm{Kamei}, \binits{Y.}},
\bauthor{\bsnm{Hassan}, \binits{A.E.}}:
\batitle{The impact of feature importance methods on the interpretation of
  defect classifiers}.
\bjtitle{IEEE Transactions on Software Engineering}
\bvolume{48}(\bissue{7}),
\bfpage{2245}--\blpage{2261}
(\byear{2022})
{\href{https://arxiv.org/abs/2202.02389}{{arXiv:2202.02389}}}
{[cs]}
\end{barticle}
\endbibitem

\bibitem[\protect\citeauthoryear{Slack et~al.}{2021}]{slackReliablePostHoc2021}
\begin{botherref}
\oauthor{\bsnm{Slack}, \binits{D.}},
\oauthor{\bsnm{Hilgard}, \binits{S.}},
\oauthor{\bsnm{Singh}, \binits{S.}},
\oauthor{\bsnm{Lakkaraju}, \binits{H.}}:
Reliable {{Post}} Hoc {{Explanations}}: {{Modeling Uncertainty}} in
  {{Explainability}}.
{arXiv}
(2021)
\end{botherref}
\endbibitem

\bibitem[\protect\citeauthoryear{Petsiuk
  et~al.}{2018}]{petsiukRISERandomizedInput2018}
\begin{botherref}
\oauthor{\bsnm{Petsiuk}, \binits{V.}},
\oauthor{\bsnm{Das}, \binits{A.}},
\oauthor{\bsnm{Saenko}, \binits{K.}}:
{{RISE}}: {{Randomized Input Sampling}} for {{Explanation}} of {{Black-box
  Models}}.
{arXiv}
(2018)
\end{botherref}
\endbibitem

\bibitem[\protect\citeauthoryear{Bodria
  et~al.}{}]{bodriaBenchmarkingSurveyExplanation2023a}
\begin{botherref}
\oauthor{\bsnm{Bodria}, \binits{F.}},
\oauthor{\bsnm{Giannotti}, \binits{F.}},
\oauthor{\bsnm{Guidotti}, \binits{R.}},
\oauthor{\bsnm{Naretto}, \binits{F.}},
\oauthor{\bsnm{Pedreschi}, \binits{D.}},
\oauthor{\bsnm{Rinzivillo}, \binits{S.}}:
Benchmarking and survey of explanation methods for black box models
\textbf{37}(5),
1719--1778
\end{botherref}
\endbibitem

\bibitem[\protect\citeauthoryear{Liu
  et~al.}{2021a}]{liuSyntheticBenchmarksScientific2021}
\begin{botherref}
\oauthor{\bsnm{Liu}, \binits{Y.}},
\oauthor{\bsnm{Khandagale}, \binits{S.}},
\oauthor{\bsnm{White}, \binits{C.}},
\oauthor{\bsnm{Neiswanger}, \binits{W.}}:
Synthetic {{Benchmarks}} for {{Scientific Research}} in {{Explainable Machine
  Learning}}.
{arXiv}
(2021)
\end{botherref}
\endbibitem

\bibitem[\protect\citeauthoryear{Liu et~al.}{2021b}]{liu2021synthetic}
\begin{botherref}
\oauthor{\bsnm{Liu}, \binits{Y.}},
\oauthor{\bsnm{Khandagale}, \binits{S.}},
\oauthor{\bsnm{White}, \binits{C.}},
\oauthor{\bsnm{Neiswanger}, \binits{W.}}:
Synthetic benchmarks for scientific research in explainable machine learning.
arXiv preprint arXiv:2106.12543
(2021)
\end{botherref}
\endbibitem

\bibitem[\protect\citeauthoryear{Han
  et~al.}{2022}]{hanWhichExplanationShould2022}
\begin{botherref}
\oauthor{\bsnm{Han}, \binits{T.}},
\oauthor{\bsnm{Srinivas}, \binits{S.}},
\oauthor{\bsnm{Lakkaraju}, \binits{H.}}:
Which {{Explanation Should I Choose}}? {{A Function Approximation Perspective}}
  to {{Characterizing Post}} Hoc {{Explanations}}.
{arXiv}
(2022)
\end{botherref}
\endbibitem

\bibitem[\protect\citeauthoryear{Turbé
  et~al.}{}]{turbeEvaluationPosthocInterpretability2023}
\begin{botherref}
\oauthor{\bsnm{Turbé}, \binits{H.}},
\oauthor{\bsnm{Bjelogrlic}, \binits{M.}},
\oauthor{\bsnm{Lovis}, \binits{C.}},
\oauthor{\bsnm{Mengaldo}, \binits{G.}}:
Evaluation of post-hoc interpretability methods in time-series classification
\textbf{5}(3),
250--260
\end{botherref}
\endbibitem

\bibitem[\protect\citeauthoryear{Ismail
  et~al.}{}]{ismailBenchmarkingDeepLearning2020}
\begin{botherref}
\oauthor{\bsnm{Ismail}, \binits{A.A.}},
\oauthor{\bsnm{Gunady}, \binits{M.}},
\oauthor{\bsnm{Corrada~Bravo}, \binits{H.}},
\oauthor{\bsnm{Feizi}, \binits{S.}}:
Benchmarking {{Deep Learning Interpretability}} in {{Time Series Predictions}}.
In: Advances in {{Neural Information Processing Systems}},
vol. 33,
pp. 6441--6452.
{Curran Associates, Inc.}
\end{botherref}
\endbibitem

\bibitem[\protect\citeauthoryear{Yang and Kim}{2019}]{yang2019benchmarking}
\begin{botherref}
\oauthor{\bsnm{Yang}, \binits{M.}},
\oauthor{\bsnm{Kim}, \binits{B.}}:
Benchmarking attribution methods with relative feature importance.
arXiv preprint arXiv:1907.09701
(2019)
\end{botherref}
\endbibitem

\bibitem[\protect\citeauthoryear{Zhong et~al.}{2023}]{zhong2023clock}
\begin{botherref}
\oauthor{\bsnm{Zhong}, \binits{Z.}},
\oauthor{\bsnm{Liu}, \binits{Z.}},
\oauthor{\bsnm{Tegmark}, \binits{M.}},
\oauthor{\bsnm{Andreas}, \binits{J.}}:
The clock and the pizza: Two stories in mechanistic explanation of neural
  networks.
arXiv preprint arXiv:2306.17844
(2023)
\end{botherref}
\endbibitem

\bibitem[\protect\citeauthoryear{Han et~al.}{2022}]{han2022explanation}
\begin{barticle}
\bauthor{\bsnm{Han}, \binits{T.}},
\bauthor{\bsnm{Srinivas}, \binits{S.}},
\bauthor{\bsnm{Lakkaraju}, \binits{H.}}:
\batitle{Which explanation should i choose? a function approximation
  perspective to characterizing post hoc explanations}.
\bjtitle{Advances in Neural Information Processing Systems}
\bvolume{35},
\bfpage{5256}--\blpage{5268}
(\byear{2022})
\end{barticle}
\endbibitem

\bibitem[\protect\citeauthoryear{Agarwal
  et~al.}{}]{agarwalOpenXAITransparentEvaluation}
\begin{botherref}
\oauthor{\bsnm{Agarwal}, \binits{C.}},
\oauthor{\bsnm{Krishna}, \binits{S.}},
\oauthor{\bsnm{Saxena}, \binits{E.}},
\oauthor{\bsnm{Pawelczyk}, \binits{M.}},
\oauthor{\bsnm{Johnson}, \binits{N.}},
\oauthor{\bsnm{Zitnik}, \binits{M.}},
\oauthor{\bsnm{Lakkaraju}, \binits{H.}}:
{{OpenXAI}}: {{Towards}} a {{Transparent Evaluation}} of {{Post}} hoc {{Model
  Explanations}},
16
\end{botherref}
\endbibitem

\bibitem[\protect\citeauthoryear{Kokhlikyan
  et~al.}{2020}]{kokhlikyanCaptumUnifiedGeneric2020}
\begin{botherref}
\oauthor{\bsnm{Kokhlikyan}, \binits{N.}},
\oauthor{\bsnm{Miglani}, \binits{V.}},
\oauthor{\bsnm{Martin}, \binits{M.}},
\oauthor{\bsnm{Wang}, \binits{E.}},
\oauthor{\bsnm{Alsallakh}, \binits{B.}},
\oauthor{\bsnm{Reynolds}, \binits{J.}},
\oauthor{\bsnm{Melnikov}, \binits{A.}},
\oauthor{\bsnm{Kliushkina}, \binits{N.}},
\oauthor{\bsnm{Araya}, \binits{C.}},
\oauthor{\bsnm{Yan}, \binits{S.}},
\oauthor{\bsnm{{Reblitz-Richardson}}, \binits{O.}}:
Captum: {{A}} Unified and Generic Model Interpretability Library for
  {{PyTorch}}.
{arXiv}
(2020)
\end{botherref}
\endbibitem

\bibitem[\protect\citeauthoryear{Hedstr{\"o}m
  et~al.}{2022}]{hedstromQuantusExplainableAI2022}
\begin{botherref}
\oauthor{\bsnm{Hedstr{\"o}m}, \binits{A.}},
\oauthor{\bsnm{Weber}, \binits{L.}},
\oauthor{\bsnm{Bareeva}, \binits{D.}},
\oauthor{\bsnm{Motzkus}, \binits{F.}},
\oauthor{\bsnm{Samek}, \binits{W.}},
\oauthor{\bsnm{Lapuschkin}, \binits{S.}},
\oauthor{\bsnm{H{\"o}hne}, \binits{M.M.-C.}}:
Quantus: {{An Explainable AI Toolkit}} for {{Responsible Evaluation}} of
  {{Neural Network Explanations}}.
{arXiv}
(2022)
\end{botherref}
\endbibitem

\bibitem[\protect\citeauthoryear{Guidotti}{2021}]{guidotti2021evaluating}
\begin{barticle}
\bauthor{\bsnm{Guidotti}, \binits{R.}}:
\batitle{Evaluating local explanation methods on ground truth}.
\bjtitle{Artificial Intelligence}
\bvolume{291},
\bfpage{103428}
(\byear{2021})
\end{barticle}
\endbibitem

\bibitem[\protect\citeauthoryear{Le et~al.}{2023}]{le2023benchmarking}
\begin{bchapter}
\bauthor{\bsnm{Le}, \binits{P.Q.}},
\bauthor{\bsnm{Nauta}, \binits{M.}},
\bauthor{\bsnm{Van Bach~Nguyen}, \binits{S.P.}},
\bauthor{\bsnm{Schl{\"o}tterer}, \binits{J.}},
\bauthor{\bsnm{Seifert}, \binits{C.}}:
\bctitle{Benchmarking explainable ai-a survey on available toolkits and open
  challenges}.
In: \bbtitle{International Joint Conference on Artificial Intelligence}
(\byear{2023})
\end{bchapter}
\endbibitem

\bibitem[\protect\citeauthoryear{Turb{\'e} et~al.}{2023}]{turbe2023evaluation}
\begin{barticle}
\bauthor{\bsnm{Turb{\'e}}, \binits{H.}},
\bauthor{\bsnm{Bjelogrlic}, \binits{M.}},
\bauthor{\bsnm{Lovis}, \binits{C.}},
\bauthor{\bsnm{Mengaldo}, \binits{G.}}:
\batitle{Evaluation of post-hoc interpretability methods in time-series
  classification}.
\bjtitle{Nature Machine Intelligence}
\bvolume{5}(\bissue{3}),
\bfpage{250}--\blpage{260}
(\byear{2023})
\end{barticle}
\endbibitem

\bibitem[\protect\citeauthoryear{Dua and Graff}{2017}]{Dua:2019}
\begin{botherref}
\oauthor{\bsnm{Dua}, \binits{D.}},
\oauthor{\bsnm{Graff}, \binits{C.}}:
{UCI} Machine Learning Repository
(2017).
\url{http://archive.ics.uci.edu/ml}
\end{botherref}
\endbibitem

\end{thebibliography}

\end{document}